\documentclass[review]{elsarticle}
\usepackage{geometry}
\usepackage{hyperref}
\usepackage[numbers]{natbib}
\usepackage{url}            
\usepackage{paralist}
\usepackage{booktabs}       
\usepackage{amsfonts}       
\usepackage{nicefrac}       
\usepackage{microtype}      
\usepackage{graphicx,subfigure,color} 
\usepackage{enumitem}
\usepackage{algorithmic,algorithm,bm}
\usepackage{lipsum}
\usepackage{amsmath,mathrsfs,amssymb,hyperref}
\usepackage{caption2}
\usepackage{dsfont}
\usepackage{multirow}
\usepackage{booktabs}       
\usepackage{standalone}     
\usepackage{preview}
\usepackage{appendix}
\usepackage{xcolor}
\allowdisplaybreaks

\usepackage{algorithm,algorithmic}

\usepackage{amsthm}
\newtheorem{myThm}{Theorem}

\newtheorem{myLemma}{Lemma}

\theoremstyle{definition}

\newtheorem{myRemark}{Remark}

\def \E {\mathbb{E}}
\def \x {\mathbf{x}}
\def \y {\mathbf{y}}
\def \yh {\hat{\y}}

\def \H {\mathcal{H}}
\def \w {\mathbf{w}}
\def \R {\mathbb{R}}

\def \F {\mathcal{F}}
\def \G {\mathcal{G}}

\def \A {\mathcal{A}}

\def \xh {\widehat{\x}}

\def \F {\mathcal{F}}

\def \X {\mathcal{X}}
\def \Xh {\hat{\X}}
\def \Y {\mathcal{Y}}
\def \Yh {\hat{\Y}}

\def \ind {\mathds{1}}
\def \O {\mathcal{O}}

\usepackage{mathtools}
\let\norm\undefined 
\DeclarePairedDelimiter\norm{\lVert}{\rVert}
\DeclarePairedDelimiter\abs{\lvert}{\rvert}

\DeclareMathOperator*{\argmin}{arg\,min}
\renewcommand{\tilde}{\widetilde}
\renewcommand{\hat}{\widehat}

\definecolor{DSgray}{cmyk}{0,1,0,0}

\DeclareFixedFont{\myfontb}{OT1}{ptm}{bx}{n}{10pt}

\def\endenv{\hfill\raisebox{1pt}{$\P$}\smallskip}
\usepackage{epstopdf}
\journal{Artificial Intelligence}

\begin{document}
\begin{frontmatter}
\title{Exploratory Machine Learning with Unknown Unknowns}
\author{Peng Zhao, Jia-Wei Shan, Yu-Jie Zhang, Zhi-Hua Zhou\corref{cor1}}
\address{National Key Laboratory for Novel Software Technology\\
Nanjing University, Nanjing 210023, China}
\cortext[cor1]{\small Corresponding author. Email: zhouzh@lamda.nju.edu.cn}

\begin{abstract}
In conventional supervised learning, a training dataset is given with ground-truth labels from a known label set, and the learned model will classify unseen instances to known labels. This paper studies a new problem setting in which there are unknown classes in the training data misperceived as other labels, and thus their existence appears unknown from the given supervision. We attribute the \emph{unknown unknowns} to the fact that the training dataset is badly advised by the incompletely perceived label space due to the insufficient feature information. To this end, we propose the \emph{exploratory machine learning}, which examines and investigates  training data by actively augmenting the feature space to discover potentially hidden classes. Our method consists of three ingredients including rejection model, feature exploration, and model cascade. We provide theoretical analysis to justify its superiority, and validate the effectiveness on both synthetic and real datasets.
\end{abstract}
\end{frontmatter}

\section{Introduction}
\label{sec:intro}
\begin{figure}[!t]
\centering
\includegraphics[width=0.9\textwidth]{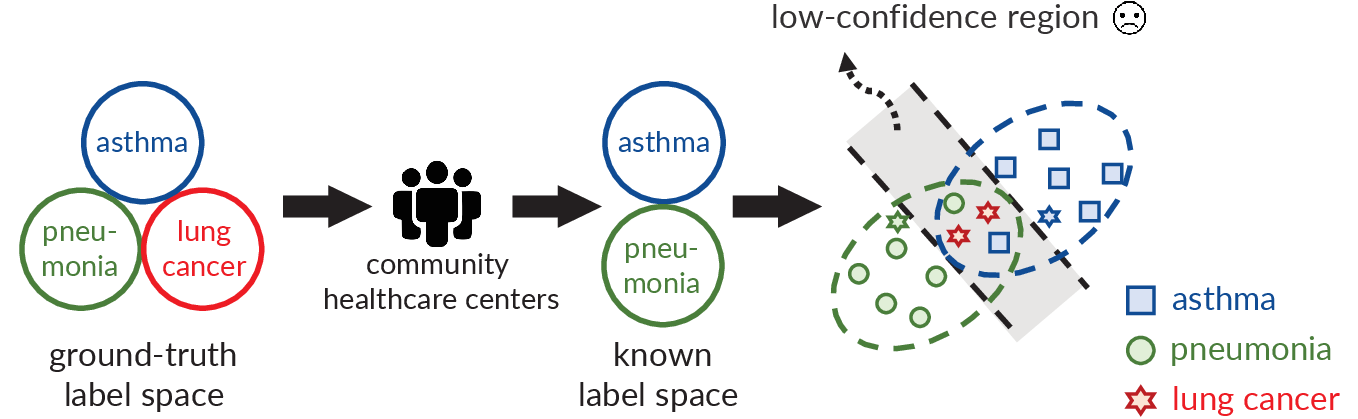}
\caption{Unknown unknowns in the task of medical diagnosis. Patients with \emph{lung cancer} are misdiagnosed as \emph{asthma} or \emph{pneumonia} due to the lack of CT scan devices, and thus appear as unknown to the learned model.}
\label{fig:example}
\end{figure}

In this paper, we study the task in which there are unknown labels hidden in the training dataset, namely some training instances belonging to a certain class are wrongly perceived as others, and thus appear unknown to the learned model. This is always the case when the label space is misspecified due to the insufficient feature information. Consider the task of medical diagnosis, where we need to train a machine learning model for the community healthcare centers based on their patient records, to help diagnose the cause of a patient with cough and dyspnea. As shown in Figure~\ref{fig:example}, there are actually three causes: two common ones (\emph{asthma} and \emph{pneumonia}), as well as an unusual one (\emph{lung cancer}). Note that the diagnosis of lung cancer crucially relies on the computerized tomography (CT) scan device, yet is  too expensive to purchase. Thus, the community healthcare centers are not likely to diagnose patients with dyspepsia as cancer, resulting in that the class of ``lung cancer'' becomes invisible and hidden in the collected training dataset. As a result, the learned model will be unaware of this unobserved class, hence facing  the unknown unknowns.

Similar phenomena occur in many other applications. For instance, the trace of a new-type aircraft was mislabeled as old-type aircrafts until performance of the aircraft detector is found poor (i.e., the capability of collected signals is inadequate), and the officer suspects that there are new-type aircrafts unknown previously. When the feature information is insufficient, there is a high risk to misperceive some classes of training data as others, leading to the existence of hidden unknown classes. More importantly, the hidden classes are sometimes of more interest, like in the above two examples. It is therefore crucial for the learned model to discover hidden classes and classify known classes well simultaneously, and this is also one of the key requirements of robust and open-world/open environment artificial intelligence~\citep{AAAI'08:open-world-AI,TGD:robust-AI,nsr'22:Open-Survey}.

The \emph{conventional supervised learning (SL)}, where a predictive model is trained on a given labeled dataset and then deployed to classify unseen instances into known labels, crucially relies on a high-quality training dataset. Thus, when the aforementioned \emph{unknown unknowns} emerged in the training data, the conventional supervised learning cannot obtain a satisfied learned model. \emph{Open category learning} (also known as \emph{learning with new classes}), which focuses on handling unknown classes appearing only in the testing phase~\citep{journals/pami/ScheirerRSB13,journals/pami/ScheirerJB14,conf/aaai/DaYZ14,conf/icml/LiuGDFH18,arXiv'19:LAC}, assumes that the unknown classes only appear in the testing stage, while in above examples there exist unknown classes in training data (see Section~\ref{sec:related-work} for more details). Neither of the learning frameworks could deal with the raised scenarios. As a result, it is necessary to develop new learning framework to handle such unknown unknowns that might emerge in the training data. 
\section{ExML: A New Learning Framework}
\label{sec:ExML}
The problem we are concerned with is essentially a class of \emph{unknown unknowns}.
In fact, how to deal with unknown unknowns is the fundamental question of robust artificial intelligence~\citep{TGD:robust-AI} and open-environment machine learning~\citep{nsr'22:Open-Survey,AAAI'20:open-world-learning}, and many studies have been devoted to addressing various aspects including changing distributions~\citep{journals/tkde/PanY10,journals/csur/GamaZBPB14,JMLR:sword++}, evolvable features~\citep{NIPS'17:FSEL,PAMI'18:houchenping,ICML'20:FDESL}, open categories~\citep{journals/pami/ScheirerRSB13,arXiv'18:open-set-Huang,NeurIPS'20:Eulac}, etc. Different from them, we study a new problem setting ignored previously, that is, the training dataset is badly advised by the \emph{incompletely perceived label space} due to the \emph{insufficient feature information}. This problem turns out to be quite challenging, since feature space and label space are entangled and \emph{both} of them are unreliable. 

Notably, it is infeasible to merely pick out instances with low predictive confidence as hidden classes, because we can hardly distinguish: (i) instances from hidden classes that suffer from low-confidence predictions owing to the incomplete label space; (ii) instances from known classes that suffer from low-confidence predictions because of insufficient feature information. This characteristic reflects intrinsic hardness of learning with unknown unknowns due to feature deficiency, and it is therefore necessary to ask for external feature information. 

There are lines of works sharing similar spirits, that is, asking for external feature information to enhance model performance, such as \emph{detecting high-confidence false predictions}~\citep{journals/jqid/AttenbergPF15,conf/aaai/LakkarajuKCH17,conf/aaai/BansalW18}, \emph{avoiding negative side effects}~\citep{IJCAI'20:NSE,AImagazine'21:NSE} and \emph{active learning}~\citep{book'12:active-learning}. However, our setting and developed methodologies are significantly different from theirs; see Section~\ref{sec:related-work} for more details. In fact, these studies as well as our work both align with the \emph{human-in-the-loop learning} principle, which leverages human knowledge to advance machine learning~\citep{JAIR'19:hitl, FGCS'22:hitl}. We believe there is potential for mutual benefit between ExML and other human-in-the-loop learning techniques, such as large language models~(LLM) trained through \emph{reinforcement learning from human feedback~(RLHF)~\citep{nips'22:rlhf}}.

\begin{figure}[!t]
\centering
\includegraphics[width=0.75\textwidth]{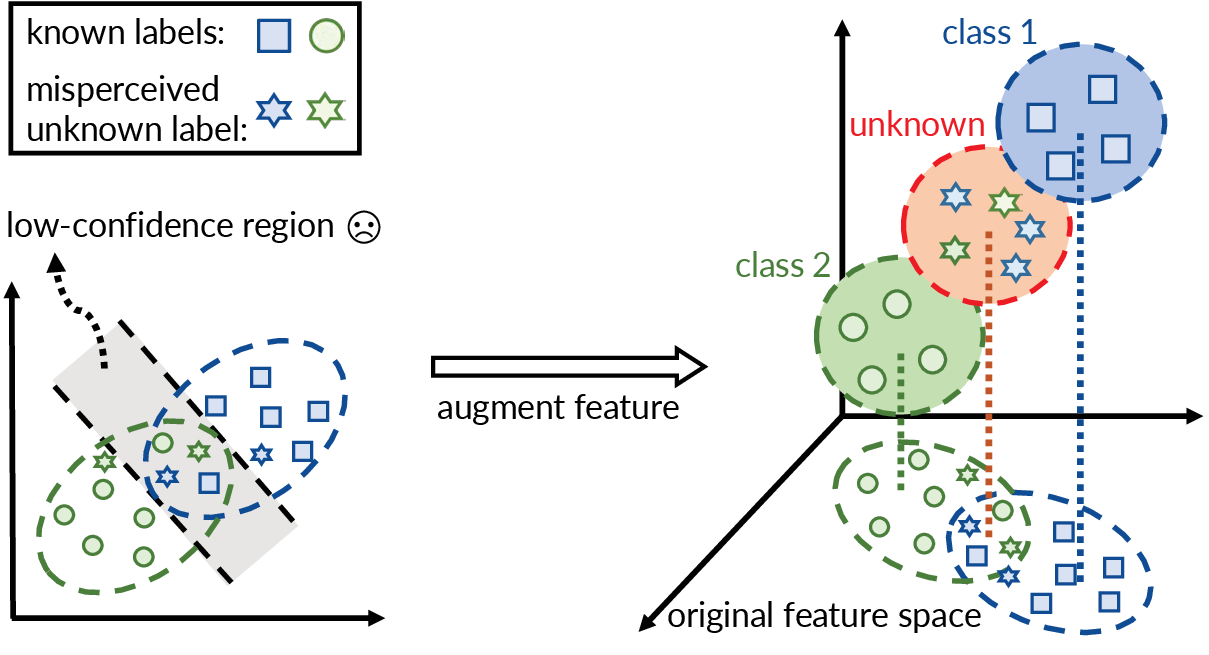}
\caption{An example illustrates that an informative feature can substantially improve  separability of low-confidence samples and make the hidden class distinguishable.}
\label{fig:augmentation}
\end{figure}

\subsection{Exploratory Machine Learning}
To handle unknown unknowns caused by the feature deficiency, we resort to the human in the learning loop to interact with environments for enhancing the data collection, more specifically, actively augmenting the feature space. The idea is that when a learned model remains performing poorly even fed with much more data, the learner will suspect existence of hidden classes and subsequently seek several candidate features to augment. Figure~\ref{fig:augmentation} shows a straightforward example that the learner receives a dataset and observes that there are two classes with poor separability, resulting in a noticeable low-confidence region. After a proper feature augmentation, the learner will then realize that there exists an additional class hidden in the training data previously due to the feature deficiency.

\begin{figure}[!t]
\centering
\includegraphics[width=0.75\textwidth]{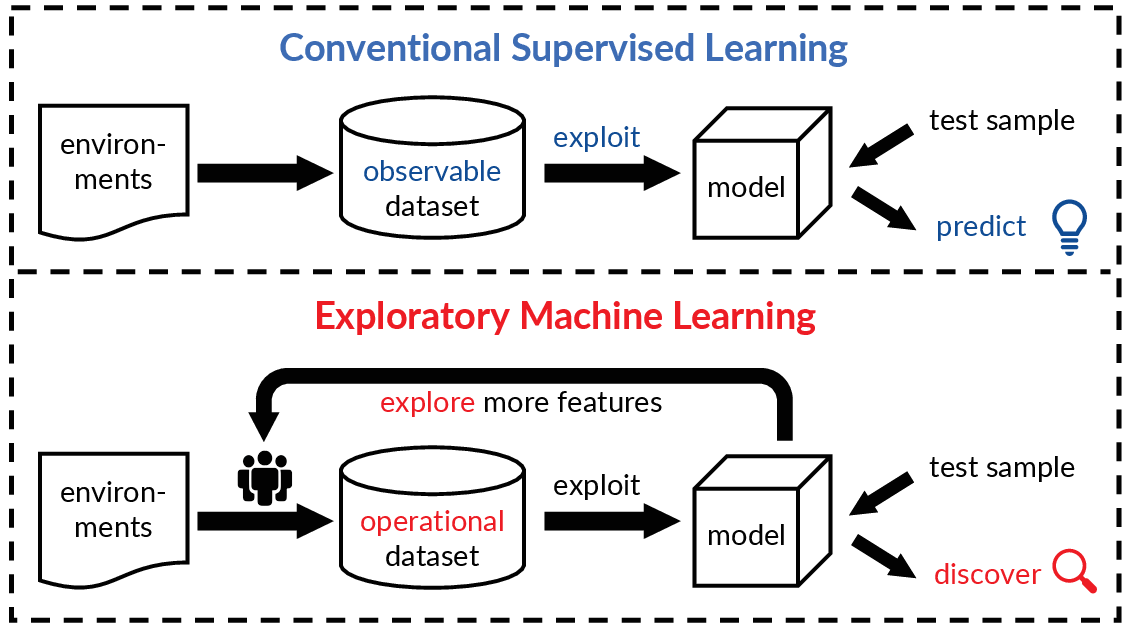}
\caption{Comparison of two learning frameworks. Conventional supervised learning exploits the observable dataset for prediction. Exploratory machine learning explores more features based on the operational dataset for both prediction and discovery of the hidden classes.}
\label{fig:comparison}
\end{figure}

Enlightened by the above example, we introduce a new learning framework called \emph{exploratory machine learning} (ExML), which explores more feature information to deal with unknown unknowns caused by feature deficiency. The terminology of exploratory learning is originally raised in the area of education, defined as an approach to teaching and learning that encourages learners to examine and investigate new material with the purpose of discovering relationships between existing background knowledge and unfamiliar content and concepts~\citep{1993:exploratory,spector2014handbook}. In the context of machine learning, our proposed framework encourages learners to \emph{examine and investigate the training dataset via exploring new feature information, with the purpose of classifying known classes and discovering potentially hidden classes}. Our proposed framework is also inspired by recent advances in cognitive science. For instance, when facing uncertain and constantly changing environments, the prefrontal cortex continuously constructs new strategies through exploration and evaluates their reliability~\citep{Science'14:cognitive-science,Science'15:cognitive-science}. Figure~\ref{fig:comparison} compares the proposed ExML to conventional supervised learning (SL). Conventional SL views the training dataset as an observable representation of environments and exploits it to train a model to predict the label. By contrast, ExML considers the training dataset is \emph{operational}, where learners can examine and investigate the dataset by \emph{exploring} more feature information, and thereby \emph{discover} unknown unknowns due to feature deficiency. Note that we does not assume that the new class necessary exists. When there is no unknown classes, our approach still offers a powerful tool to present feature exploration to help refine the performance of conventional supervised learning. 

We further develop an approach to implement the principle of ExML, consisting of three important ingredients: rejection model, feature exploration, and model cascade. The rejection model identifies suspicious instances that potentially belong to the hidden classes. Feature exploration guides which feature should be explored among the candidates, and then retrains the model on the augmented feature space. Model cascade allows a layer-by-layer processing to refine the selection of suspicious instances. Theoretical analysis is provided to justify the superiority of our proposed framework. Besides, we present empirical evaluations on synthetic data to illustrate the idea and further validate the effectiveness on real datasets.

\subsection{Problem Formulation}
\label{sec:problem-formulation}

\paragraph{Training Dataset} The learner receives a training dataset $\hat{D}_{tr} = \{(\xh_i,\yh_i)\}_{i=1}^m$, where the feature $\xh_i \in \hat{\X} \subseteq \R^d$ is from the \emph{observed} feature space and the label $\yh_i \in \Yh$ is from the \emph{incomplete} label space with $N$ known classes. Throughout the paper, we focus on the binary case for simplicity. We remind that in our concerned unknown unknowns setting there exist training samples that are actually from hidden classes yet wrongly labeled as known classes due to feature deficiency. 

\paragraph{Candidate Features and Cost Budget} Besides the training dataset, the learner can access a set of candidate features $\mathcal{A}=\{a_1, \ldots, a_K\}$, whose values are \emph{unknown} before acquisition. Moreover, a certain cost $c_i$ will be incurred to acquire an observation on the candidate feature $a_i$ for any sample. The learner aims to identify top $k$ informative features from the pool under the given budget $B > 0$ such that she will then augment the dataset on those top informative features in the testing stage. For convenience, we focus on the case that the learner desires to find the best feature, i.e., $k=1$.

We expand the two examples in the introduction to demonstrate the rationality of our formulation. In the first example, suppose a patient's physical examination results suggest that he might have pneumonia, but the diagnosis is at a low confidence. At this point, the doctor may recommend the patient to do further examinations (i.e., the pulmonary histopathology examination, the CT scans, etc.) which can be regarded as candidate features in our setting. The cost of doing these examinations varies, and the assistance they may provide for a more accurate diagnosis also differs. In the second example, when the performance of the aircraft is found poor, the detector may ask for more sources of signals (i.e., optical sensors, aviation sonar, etc.). The signals generated by the new equipment can be regarded as candidate features in our setting. The cost of deploying these devices varies, and the effectiveness of the signals also differs.

\paragraph{Testing Stage} Suppose the learner identifies the best feature as $a_i$, she will then augment the testing sample with this particular feature in the feature space, leading to an augmented feature space denoted by $\X_i = (\hat{\X}\cup\X^i)\subseteq \R^{d+1}$ where $\X^i$ is the feature space of $a_i$ and recall that $\hat{\X} \in \R^d$ is the original feature space. The learned model requires predicting the label of the augmented testing sample, either classified to one of known classes or discovered as hidden classes (abbrev. \textsf{hc}).

\begin{myRemark}[Possible relaxations of some assumptions]
  We have made several modeling assumptions are introduced in the above problem formulation for simplicity, with the aim of avoiding distractions of an over-complicated setting and better understanding the essence of this new problem setup. Indeed, our proposed principle can still work when relaxing these assumptions by borrowing more advanced techniques. For example, we can leverage multi-class rejection techniques~\citep{journals/JASA/ZhangWX2018,Neurips'19:multiclass-rejection} to generalize our framework into multi-class problems, and use top-$k$ best arm identification~\citep{ICML'12:BAI-top-k,AISTAS'17:BAI-top-k-instance} to select multiple augmented features. We leave those potential extensions as future works.
  \endenv
\end{myRemark}
\begin{myRemark}[Training-time and test-time feature cost]
  Our problem formulation captures the training-time feature cost, which means the learner is required to pay for acquiring new features for the training samples. Note that in the testing stage, augmenting the testing sample with candidate features may also incur a certain cost. Our paper focuses on the training-time feature cost and designs budget allocation strategies for feature exploration, while it is also possible to extend our framework to further accommodate test-time feature cost by modifying the goal of feature exploration, for example, to encourage the algorithm to identify the feature with highest quality-cost ratio~\citep{journals/infocom/qin2020exploring}. We leave the extension to test-time feature cost as future work. 
  \endenv
\end{myRemark}

\section{A Practical Approach}
\label{sec:approach}
 Due to the feature deficiency, the learner might be even unaware of the existence of hidden classes based on the observed training data. It is thus necessary to introduce the assumption that \emph{instances with high predictive confidence are safe, i.e., they will be correctly predicted as one of the known classes}. The learner will suspect the existence of hidden classes (which are the unknown unknowns to the learner at the beginning) when the learned model performs badly.

We justify the necessity of the assumption. Actually, there are some previous works studying the problem of high-confidence false predictions without considering the issue of feature deficiency~\citep{journals/jqid/AttenbergPF15,conf/aaai/LakkarajuKCH17}, in which there exist some instances that are wrongly predicted with high confidence. Since the model's performance is highly unreliable, to rectify that, they assume the existence of an oracle providing ground-truth labels for the given queries.
However, in the presence of the feature deficiency as in our scenario, the problem would not be tractable unless there is an oracle able to provide ground-truth labels based on the insufficient feature representation, which turns out to be an even stronger assumption that does not hold in reality generally. So this paper focuses on the aforementioned case to trust the high-confidence predictions and we leave high-confidence unknown unknowns due to the insufficient feature as the future work to explore. 

We further clarify and emphasize that the introduced assumption does not trivialize the problem setup, because notice that the low-predictive instances are \emph{not} necessarily from hidden classes (as explained at the beginning of Section~\ref{sec:ExML}), which necessitates more efforts in discovering and identifying unknown unknowns. Following the methodology of ExML (examining the training dataset via exploring new feature information), we design a novel approach, which consists of three components: rejection model, feature exploration, and model cascade. Figure~\ref{fig:architecture} illustrates the main procedures, and we will describe the details of each component subsequently.

\subsection{Rejection Model}
\label{sec:reject-model}
\begin{figure}[!t]
\centering
  \begin{subfigure}[overall procedure]{
    \begin{minipage}[d]{0.46\textwidth}
    \label{fig:syn-allocation-1}
     \includegraphics[clip, trim=0.2cm 0cm 13.12cm 0cm,width=\textwidth]{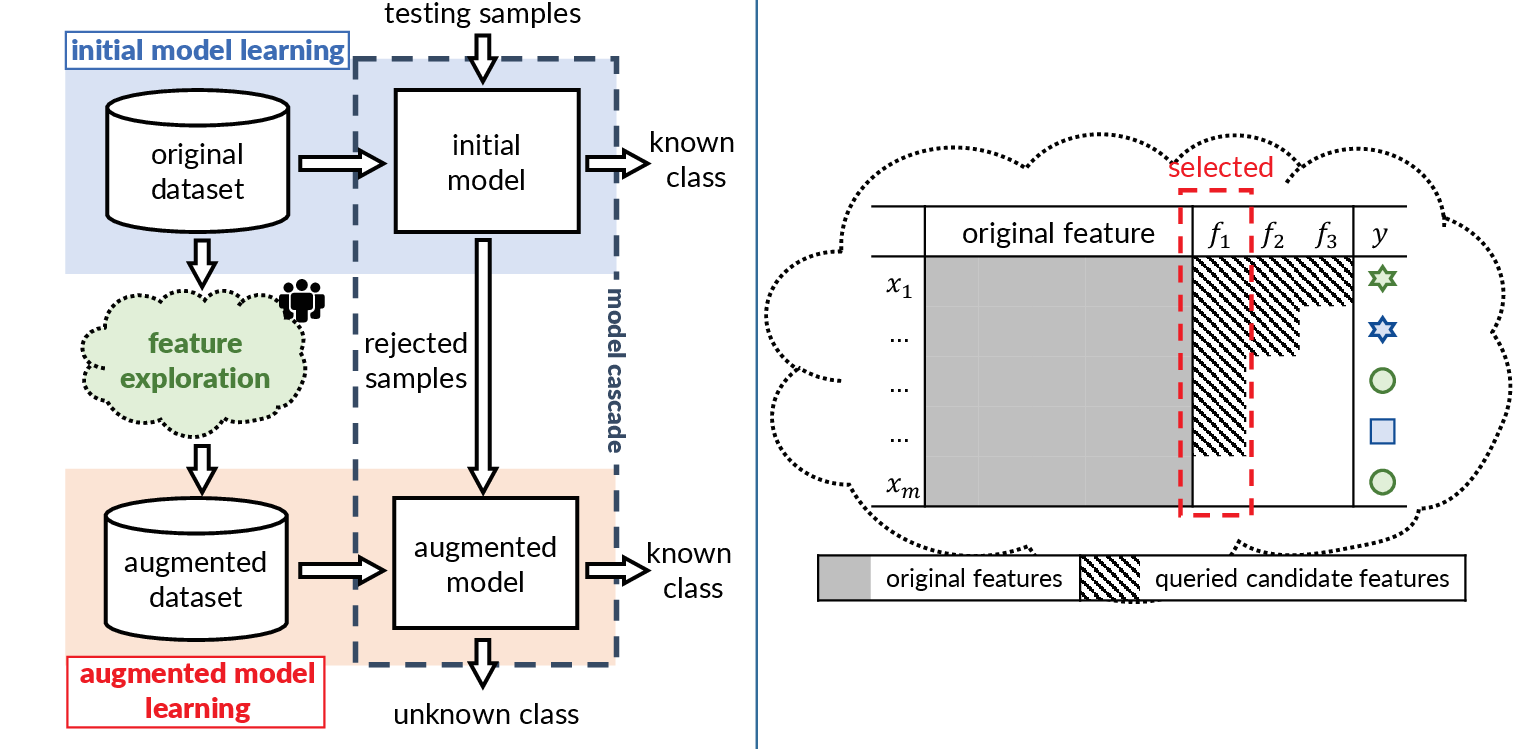}
     \end{minipage}}
  \end{subfigure}
  \begin{subfigure}[feature exploration]{
    \begin{minipage}[d]{0.46\textwidth}
    \label{fig:syn-allocation-2}
     \includegraphics[clip, trim=12.92cm 0cm 0.05cm 0cm,width=\textwidth]{figure/architecture_v4}
     \end{minipage}}
  \end{subfigure}
  \caption{The left figure shows the overall procedure of ExML. Our approach begins with an initial model (blue part), followed by exploring the best candidate feature among the candidates (green part). Afterwards, a learned model is retrained based on the augmented dataset, and finally is cascaded with the initial model to discover the hidden class (red part). The right figure describes the procedure of the feature exploration in ExML.}
  \label{fig:architecture}
\end{figure}

As shown in Figure~\ref{fig:syn-allocation-1}, at the beginning, the learner requires to train an initial model on the original dataset, with capability of identifying low-confidence instances. As emphasized previously (cf. the beginning of Section~\ref{sec:ExML}), these low-confidence instances could come from both known and hidden classes, so they are only detected as suspicious and will be refined in the further procedures.

In order to obtain such models, we leverage the techniques of learning with rejection~\citep{ALT16:reject-theory}, where the learned model will abstain from predicting instances whose maximum conditional probabilities are lower than a given threshold. More precisely, we learn a function pair $f=(h,g)$, where $h:\hat{\X} \mapsto \R$ is the \emph{predictive} function for the known classes and $g:\hat{\mathcal{X}}\mapsto\mathbb{R}$ is the gate function to \emph{reject} the hidden class. The sample $\hat{\x}$ is classified to the hidden class if $g(\hat{\x})<0$, and otherwise to the class of $\mathtt{sign}(h(\hat{\x}))$. Such rejection models can be trained via optimizing the following objective:
\begin{equation}
\label{eq:h-g}
      \min\nolimits_{f}\ \mathbb{E}_{(\xh,\yh)\sim\hat{\mathcal{D}}}[\ell_{0/1}(f,\xh,\yh;\theta)],
\end{equation}
where $$\ell_{0/1}(f,\xh,\yh;\theta) = \ind_{\yh\cdot h(\hat{\mathbf{x}})<0}\cdot  \ind_{g(\hat{\mathbf{x}})>0} + \theta\cdot  \ind_{g(\hat{\mathbf{x}})\leq0}$$ is the 0-1 loss of the rejection model $f$ parameterized by the threshold $\theta \in (0,1)$ and $\hat{\mathcal{D}}$ is the data distribution over $\Xh\times\Yh$. To tackle the difficulty of non-convex optimization arising from the indicator function,~\citet{ALT16:reject-theory} introduce the following surrogate loss function
\begin{equation}
\label{eq:surrogate-loss}
 \ell_{surr}(f,\hat{\mathbf{x}},\yh;\theta) = \max \left\{1+\frac{1}{2}\big(g(\hat{\mathbf{x}})-\yh\cdot h(\hat{\mathbf{x}})\big),\theta\cdot \Big(1-\frac{g(\hat{\mathbf{x}})}{1-2\theta}\Big),0\right\}
\end{equation}
to approximate the original $\ell_{0/1}$ loss. Since the distribution is unknown and we cannot directly measure the risk, we choose the model that minimizes the empirical risk:
\begin{equation}
\label{eq:surrogate}
  \min_{f\in\mathbb{H}\times\mathbb{H}}~~ \frac{1}{m}\sum_{i=1}^{m} \ell_{surr}(f,\hat{\mathbf{x}}_i,\yh_i;\theta) + C_h \Vert h \Vert^2_{\mathbb{H}} + C_g\Vert g\Vert^2_{\mathbb{H}},
\end{equation}
where $C_h$ and $C_g$ are regularization parameters, and $\mathbb{H}$ is the RKHS induced by the kernel $K:\hat{\mathcal{X}}\times\hat{\mathcal{X}}\mapsto \mathbb{R}$. By the representer theorem~\citep{books/lib/ScholkopfS02}, the optimizer of~\eqref{eq:surrogate} is in the form of $h(\hat{\mathbf{x}}) = \sum_{i=1}^m u_i K(\hat{\mathbf{x}},\hat{\mathbf{x}}_i)$ and $g(\hat{\mathbf{x}}) = \sum_{i=1}^m w_i K(\hat{\mathbf{x}},\hat{\mathbf{x}}_i)$, where $u_i$ and $w_i$ are coefficients to learn. So~\eqref{eq:surrogate} can be reformulated as quadratic programming and solved efficiently. 

\begin{myRemark}[Reliability of the initial model]
  The reliability of the initial model is crucial to make ExML effective. Fortunately, we have many methods to enhance the reliability of the initial model. Since the training of the initial model goes as a standard process of conventional supervised learning, we can make use of any standard supervised learning techniques (e.g., data enhancement, feature engineering) to make the initial model more reliable. Besides, we can also adjust the rejection model (e.g., reduce the rejection cost $\theta$) to make it easier to meet the assumption we made at the beginning of Section~3 (instances with high predictive confidence are safe), at a cost of rejecting more samples and passing them to the subsequent models. \endenv
\end{myRemark}

\subsection{Feature Exploration}
\label{sec:feature-explore}

If the initial model is unqualified (for instance, it rejects too many samples for achieving the desired accuracy), the learner will suspect the existence of hidden classes and explore new features to augment. In our setting, the learner requires to select the best feature from $K$ candidates and retrain a model based on the augmented data, as shown in Figure~\ref{fig:syn-allocation-2}. 

We emphasize that conventional feature selection is not feasible here, because it requires to know the values of candidate features, while these values are unknown before acquisitions. To address the challenge, we propose a novel procedure---\emph{feature exploration}---to adaptively identify the most informative feature under the cost budget, \emph{without} requiring feature values in advance. To address the issue, there are two fundamental questions to answer: 
\begin{compactitem}
	\item how to measure the quality of candidate features? 
	\item how to allocate the budget to identify the best feature? 
\end{compactitem}
In the following, we will answer these two questions and then describe our strategy for the feature exploration in ExML.

\noindent\textbf{Feature quality measure.}~~ Denote by $\mathcal{D}_i$ the data distribution over $\X_i\times\Yh$, where $\X_i$ is the augmented feature space of the $i$-th candidate feature. Recall that the augmented feature space is defined as $\X_i = (\hat{\X}\cup\X^i)\subseteq \R^{d+1}$ where $\X^i$ is the feature space of $a_i$, see more notation details in Section~\ref{sec:problem-formulation}. Then, we use the \emph{Bayes risk} on $\mathcal{D}_i$ as the feature quality measure, defined as
\begin{equation}
\label{eq:criterion-bayes-risk}
  R_i^* =R_i(f^*_i) = \min\nolimits_{f}\ \mathbb{E}_{(\x,\yh)\sim\mathcal{D}_i}\big[\ell_{0/1}(f,\x,\yh;\theta)\big],
\end{equation}
where $R_i(f)$ is the expected $0/1$ risk of function $f$ over $\mathcal{D}_i$, and $f^*_i$ minimizes $R_i(f)$ over all measurable functions. The Bayes risk essentially reflects the minimal error of any rejection model that can attain on the augmented data distribution. The value will be smaller when the selected augmented feature improves the separability more significantly, and thus the associated feature is believed more informative.

Due to the inaccessibility of the underlying distribution $\mathcal{D}_i$, we approximate the Bayes risk by its empirical version evaluated on surrogate loss over the augmented data $D_i=\{(\x_j,\yh_j)\}_{j=1}^{n_i} $,
\begin{equation}
  \label{eq:criterion-bayes-risk-empirical}
  \hat{R}_{i}^{surr}(\hat{f}_i) = \frac{1}{n_i} \sum_{j=1}^{n_i} \ell_{surr}(\hat{f}_i,\x_j,\yh_j;\theta),
\end{equation}
where $\x_j\in\X_i, \yh_j\in\Yh$, and $\hat{f}_i$ is the rejection model learned by ERM over the surrogate loss~\eqref{eq:surrogate} on augmented dataset $D_i$. We prove that the approximation by surrogate loss almostly does no harm to the theoretical guarantees on the performance of the proposed algorithm (in Section~\ref{sec:theory}), and even better, we verify in experiments that the empirical surrogate loss is easy to be well-optimized to obtain an augmented feature with high quality (in Section~\ref{sec:experiment}). 

Based on the feature quality measure~\eqref{eq:criterion-bayes-risk} and its empirical version~\eqref{eq:criterion-bayes-risk-empirical}, we now introduce the budget allocation strategy to identify the best candidate feature.

\noindent\textbf{Budget allocation strategy.}~~ The goal of the feature exploration is to identify the best feature within the limited budget, and meanwhile the model retrained on augmented data should have good generalization ability. Note that the feature quality is definitely unknown to the learner. 

We first consider the simplified case of uniform cost, namely, $c_1=c_2=\dots=c_K=1$. For this setting, we propose two feature exploration strategies: uniform allocation and median elimination. Below, we describe the details.

\textbf{\textit{Uniform Allocation.}}~~ We have the uniform allocation strategy as follows, under the guidance of criterion~\eqref{eq:criterion-bayes-risk}. For each candidate feature $a_i$, $i \in [K]$, learner allocates $\lfloor B/K\rfloor$ budget and obtains an augmented dataset $D_i$. We can thus compute the empirical feature measure by~\eqref{eq:criterion-bayes-risk-empirical}, and select the feature with the smallest risk. The above strategy is simple yet effective. We prove that ExML equipped with uniform allocation as the feature exploration strategy can achieve a low excess risk with high probability, as demonstrated in Theorem~\ref{thm:uniform-allocation} of Section~\ref{sec:theory}.

\textbf{\textit{Median Elimination.}}~~ We further propose another variant inspired by the bandit theory~\citep{2019:bandit-book} to improve the budget allocation efficiency. Specifically, we adopt the technique of \emph{median elimination}~(ME)~\citep{JMLR'06:Elimination}, which removes one half of poor candidate features after every iteration and only the best one remains in the end, and proposed Algorithm~\ref{alg:median-elimination} which can avoid allocating too many budgets on poor features. More specifically, the elimination proceeds in $T = \lceil\log_2 K\rceil$ episodes, in each episode, $\lfloor B/T\rfloor$ budget is allocated uniformly to all remaining candidate features, and the learner could query their values for updating the corresponding augmented datasets $D_i$. Then, the score $\hat{R}_{i}^{surr}$ is calculated on the current augmented datasets $D_i$ and the half features with high $\hat{R}_{i}^{surr}$ are eliminated. In the last, only one candidate feature $a_{i_s}$ will be left and its augmented dataset $D_{i_s}$ contains around $\lfloor B/\log K\rfloor$ samples, which is the largest among all the candidate features.

\begin{algorithm}[!t]
   \caption{Median Elimination for Feature Exploration}
   \label{alg:median-elimination}
\begin{algorithmic}[1]
    \REQUIRE Feature exploration budget $B$, original dataset $\hat{D}_{tr} = \{(\xh_i,\yh_i)\}_{i=1}^m$, candidate feature pool $\mathcal{A} = \{a_1,\ldots,a_K\}$, threshold $\theta \in (0,1)$.
    \ENSURE Selected feature $c_{i_s} \in \mathcal{A}$ and  corresponding augmented model $\hat{f}_{i_s}$.
    \STATE Initialize: dataset $D_i = \varnothing$ for each feature $a_i\in\mathcal{A}$, set of active features $\A_1 = \mathcal{A}$, $T=\lceil\log_2 K\rceil$.
    \FOR{$t = 1,\dots,T$}
    \STATE Randomly select $n_t = \left\lfloor B/(T\vert\A_t\vert)\right\rfloor$ samples from $\hat{D}_{tr}$ and query active features $a_i\in\A_t$;
    \STATE Update $D_i$ with selected samples and train a model $\hat{f}_{t, i}$ on $D_i$ by ERM~\eqref{eq:surrogate}, for all $a_i\in\A_t$;
    \STATE Compute $\hat{R}_{t,i}^{surr}$ according to~\eqref{eq:criterion-bayes-risk-empirical}, for all $a_i \in \A_t$;
    \STATE Update $\A_{t+1}$ as half of features in $\A_t$ with lower $\hat{R}_{t,i}^{surr}$;
    \ENDFOR
\end{algorithmic}
\end{algorithm}

As shown in Figure~\ref{fig:syn-allocation-2}, poor features are eliminated earlier, the budget left for the selected feature is thus improved from $\lfloor B/K\rfloor$ to $\lfloor B/\log K\rfloor$ by Algorithm~\ref{alg:median-elimination}, which ensures better generalization ability of the learned model. The behavior is formally justified in Theorem~\ref{thm:median-elimination}. In a nutshell, we find that median elimination shows its advantage in exploring the best candidate feature more efficiently than uniform allocation despite its higher probability that fails to identify the best candidate feature than uniform allocation, since both strategies enjoy an exponentially-decayed failing probability. We finally remark that our paper currently focuses on identifying the best feature, and our framework is ready for top $k$ features identification ($k>1$) by introducing more sophisticated techniques~\citep{ICML'12:BAI-top-k,AISTAS'17:BAI-top-k-instance}. Feature exploration in our approach also shares similar ideas with a recent line of works called \emph{feature budget learning}~\citep{COLT'93:FBL,ICML'12:FBL,ICML'15:FBL} (see Section~\ref{sec:related-work} for more discussions). We believe that further leveraging the techniques from feature budget learning could be beneficial to our feature exploration problem. 

\textbf{\textit{Non-uniform Query Cost.}}~~ We have assumed so far that the query of different features shares the same cost (unit-cost setting, i.e., $c_1=c_2=\dots=c_K=1$), and now we relax this assumption by considering the more general \emph{non-uniform} cost for different candidate features, i.e., $c_1,c_2,\dots,c_K$ can be  distinct. While our goal remains as identifying the best feature within the limited budget and meanwhile obtaining good generalization ability, new consideration appears after the non-uniform cost nature that the feature exploration algorithm should balance between querying good but expensive features and querying cheap but low-quality features. As a consequence, the feature exploration phase aims to identify the best candidate feature (namely, feature $a_1$) and meanwhile to ensure that there are a large number of queries in this returned feature. To this end, we propose two principles for adapting strategies to the non-uniform case. 

\begin{itemize}
  \item \textit{Sample Alignment.}~The first one is the \emph{sample alignment principle}, where at each time we are allocating budget, the budget allocated to each active feature are aligned to ensure that a same number of samples is queried for each active feature. Specifically, when a total budget $b$ is to be allocated to a set $A$ of active features, the learner allocates to each feature $a_i\in A$ a total budget of $\big\lfloor \frac{c_ib}{\sum_{a_j\in A} c_j}\big\rfloor$ to ensure that a total number of $\big\lfloor \frac{b}{\sum_{a_j\in A} c_j} \big\rfloor$ samples are queried for each active feature. 

  \item \textit{Budget Alignment.}~We further have another variant to improve the budget allocation efficiency, which is called the \emph{budget alignment principle}. Specifically, when a total budget $b$ is to be allocated to a set $A$ of active features, the learner equally allocates to each feature a total budget of $\lfloor b/|A|\rfloor$, and thus $\lfloor b/(|A|c_i) \rfloor$ samples are queried for any active feature $a_i\in A$. Intuitively, we can have more training samples augmented with cheaper candidate feature, which possibly leads to a better generalization ability if the candidate feature has a relatively high quality. Therefore, the budget alignment principle may provide a better performance, since cheap features with relatively high quality are more sufficiently explored.
\end{itemize}

\subsection{Model Cascade}
\label{sec:model-cascade}
After the feature exploration, the learner will retrain a model on the augmented data. Considering that the augmented model might not  always be better than the initial model, particularly when the budget is not enough or candidate features are not quite informative, we employ the ensemble method by proposing the \emph{model cascade} mechanism to cascade the augmented model with the initial one. Concretely, high-confidence predictions are accepted in the initial model, the rest suspicious are passed to the next layer for feature exploration, those augmented samples with high confidence will be accepted by the augmented model, and the remaining suspicious continue to the next layer for further refinements. At the final layer, those samples with multiple refinements but are still suspicious will be classified into the unknown new class. 

Essentially, our approach can be regarded as \emph{a layer-by-layer processing for identifying instances of hidden classes}, and the procedures can be stopped until human discovers remaining suspicious are indeed with certain hidden structures. For simplicity, we only implement a two-layer architecture, that is, the suspicious samples in the second layer will be classified into the unknown new class.

Our proposed multi-layer model cascade provides a way of hierarchical refinements, but at a cost of error composition or overfitting during the learning process. Note that our model cascade strategy can be regarded as a sequential cascaded ensemble, thus, the aforementioned issues can be potentially alleviated by the techniques from \emph{ensemble learning}~\citep{book'12:ensemble-learning}. Several interesting observations can be made from the view of ensemble learning. For example, since \emph{diversity} is crucial for the success of ensemble learning~\citep{book'12:ensemble-learning,FCS18:structure-ensemble}, our proposed ExML framework may further benefit from  diversity encouragement among multiple base learners (i.e., different models in the multi-layer cascade structure), such as bagging~\citep{MLJ'96:bagging} and selective ensemble~\citep{AIJ'02:selective-ensemble}, etc. Moreover, it would be also useful to introduce diversity in the feature exploration, which is left as an interesting future work.

\section{Theoretical Analysis}
\label{sec:theory}
In this section, we present theoretical analysis for our proposed exploratory machine learning (ExML) framework. Specifically, we first investigate the attainable excess risk of supervised learning, supposing that the best feature were \emph{known} in advance. Next, we analyze the excess risk of ExML, demonstrating its effectiveness in terms of both the selection criterion and budget allocation strategies. In the following, we first present the theoretical result for supervised learning with known best feature (Section~\ref{sec:SL-theory}), and then provide the guarantee for ExML with unknown best feature (Section~\ref{sec:ExML-theory}). The proofs are deferred to~\hyperref[sec:proofs]{Appendix~A}.

Throughout the section, for each candidate feature $a_i$, we denote the corresponding hypothesis space as $\mathcal{H}_i,\mathcal{G}_i=\{\x\mapsto\langle\w,\Phi_i(\x)\rangle \mid \norm{\w}_{\mathbb{H}_i}\leq\Lambda_i\}$, where $\Phi_i$ and $\mathbb{H}_i$ are induced feature mapping and RKHS of kernel $K_i$ in the augmented feature space, and we also define $\kappa_i^2 = \sup_{\x\in\X_i}K_i(\x,\x)$. For simplicity and without loss of generality, we assume that the feature indices are sorted in ascending order based on their associated feature quality, i.e., $R_{1}^*\leq\cdots\leq R_{K}^*$. 

\subsection{Supervised Learning with Known Best Feature}
\label{sec:SL-theory}
Suppose the best feature were known in advance. Given a budget $B$ and the unit uniform cost of different features, evidently we could obtain $B$ samples augmented with this particular (best) feature $a_1$. Let $f_{\text{SL}}$ be the model learned by supervised learning via minimizing the objective~\eqref{eq:surrogate}. According to the standard learning theory literature~\citep{ALT16:reject-theory,MLSS'03:Bousquet}, we know that for any $\delta>0$, with probability at least $1-\delta$, the excess risk is bounded by
\begin{equation}
    \label{eq:sl-excess-risk}
    R_1(f_{\text{SL}})-R_1^*\leq\O\bigg(\sqrt{\frac{(\kappa_1\Lambda_1)^2}{B}}+\sqrt{\frac{\log(1/\delta)}{2B}}\bigg)+R_{ap},
\end{equation}
where $R_{ap} = \inf_{f\in\mathcal{H}_1\times\mathcal{G}_1} R^{surr}_1(f) - \inf_{f} R_{1}(f)$ is the approximation error. Note that the definition of $R_{ap}$ here is slightly more than the classical definition of approximation error that measures how well hypothesis spaces $\mathcal{H}_1$, $\mathcal{G}_1$ approach the target in terms of the expected risk $R_{1}(f)=\mathbb{E}_{(\x,\yh)\sim\mathcal{D}_1}[\ell_{0/1}(f,\x,\yh;\theta)]$ in the statistical learning literature~\citep{book'2018:foundation}, since our definition additionally counts the approximation error owing to optimizing the surrogate loss $\ell_{surr}(f)$ during the learning process instead of $\ell_{0/1}(f)$ due to the hardness of its non-convexity. Thus, if the best feature were \emph{known} in advance, the excess risk of supervised learning would converge to the inevitable approximate error in the rate of $\O(1/\sqrt{B})$, with a given budget $B$.

\subsection{Exploratory Learning with Unknown Best Feature}
\label{sec:ExML-theory}
In reality, however, \emph{the best feature is unfortunately unknown ahead of time}. More importantly, since the values of $K$ candidate features are unavailable, it is \emph{infeasible} to perform the feature selection. We show that by means of ExML (feature exploration), the excess risk also converges in a favorable rate, yet \emph{without} requiring to know the best feature in advance. Below, we first introduce a key decomposition of excess risk in generic ExML (Section~\ref{sec:ExML-decomposition}), then present the theoretical result of ExML equipped with the uniform allocation (Section~\ref{sec:ExML-theory-ua}) and ExML with median elimination (Section~\ref{sec:ExML-theory-me}), respectively.

\subsubsection{Key Decomposition and Exploratory Regret}
\label{sec:ExML-decomposition}
We first introduce the notations and an assumption used throughout the theoretical analysis in ExML, then present the key decomposition which demonstrates the different challenges in ExML comparing with conventional SL. 

\paragraph*{Notations and assumption}~We use $\hat{D}_{tr, i}$ to denote the entire training dataset augmented with feature $a_i$ and use $\hat{R}_{tr, i}^{surr}(f)$ to denote the averaged surrogate risk on $\hat{D}_{tr, i}$. Let $\hat{f}_i^*\in\H_i\times\G_i$ be the minimizer of $\hat{R}_{tr, i}^{surr}(f)$, namely, $\hat{f}_i^* \in \argmin_{f \in \H_i\times\G_i} \hat{R}_{tr, i}^{surr}(f)$. To facilitate the theoretical analysis, we introduce the assumption that \emph{the most informative feature leads to the smallest loss on the entire augmented training dataset}, more specifically, $\hat{R}_{tr, 1}^{surr}(\hat{f}_1^{*}) = \min_{i\in[K]} \hat{R}_{tr, i}^{surr}(\hat{f}_i^{*})$, noting that as mentioned earlier the features are supposed to be sorted according to the quality, $R_{1}^*\leq\cdots\leq R_{K}^*$, without loss of generality. 

The assumption is natural in the sense that when deploying ExML framework to tackle unknown unknowns, one should already have tried collecting a relatively large training dataset (but without feature augmentation), so evaluating on the empirical data should be able to reflect the underlying feature quality. Moreover, the assumption is also necessary to the best of our understanding, because suppose otherwise, the most informative feature cannot be identified through the empirical data even with an unlimited feature budget, then obviously any algorithm can hardly approach a desired excess risk. 

\begin{myRemark}[Most informative feature assumption over 0/1 loss]
  One can notice that the assumption is made on the surrogate loss, while the feature quality is measured via the 0/1 loss. In fact, the assumption is to guarantee the performance of feature exploration, which includes feature quality evaluations on surrogate loss by ERM. Therefore, the loss function in the assumption should be aligned with the loss function used in the feature exploration algorithm. However, due to the difficulty of non-convex optimization, it is generally hard to proceed ERM on the 0/1 loss, thus it remains unclear whether we can obtain the same guarantees when making such an assumption over 0/1 loss, which is an interesting future issue to explore. \endenv
\end{myRemark}

We measure the performance of ExML by the excess risk $R_{i_s}(\hat{f}_{i_s})-R_1^*$, which is the difference between the expected risk of the hypothesis $\hat{f}_{i_s}$ returned by ExML evaluated over the augmented feature space $\X_{i_s}$ and the Bayes risk $R_1^*$ over the best augmented feature space $\X_1$. To proceed the theoretical analysis, we introduce an important quantity used in analyzing the behavior of ExML algorithms, defined as 
\begin{equation}
  \label{eq:optimality_gap}
  \Delta_i=\hat{R}_{tr, i}^{surr}(\hat{f}_i^{*})-\hat{R}_{tr, 1}^{surr}(\hat{f}_1^{*}),
\end{equation}
which qualifies the empirical difference of feature quality between feature $i$ and that of the best feature. Let $\Delta=\min_{i\in[K], \Delta_i>0}\Delta_i$ be the smallest one in the candidate features, which we call as \emph{optimality gap} measuring the hardness of  feature exploration in ExML. 

The key step in the analysis of generic ExML which demonstrates the different challenges comparing to SL is to decompose the excess risk of the learned model $\hat{f}_{i_s}$ into five parts,
\begin{equation}
  \begin{aligned}
    R_{i_s}(\hat{f}_{i_s}) - R_1^* &= \underbrace{R_{i_s}(\hat{f}_{i_s}) - \hat{R}_{tr, i_s}^{surr}(\hat{f}_{i_s})}_{\mathtt{term~(a)}} + \underbrace{\hat{R}_{tr, i_s}^{surr}(\hat{f}_{i_s}) - \hat{R}_{tr, 1}^{surr}(\hat{f}_1^*)}_{\mathtt{term~(b)}} \\
    &+ \underbrace{\hat{R}_{tr, 1}^{surr}(\hat{f}_1^*) - \hat{R}_{tr, 1}^{surr}(f_1^*)}_{\mathtt{term~(c)}} + \underbrace{\hat{R}_{tr, 1}^{surr}(f_1^*) - R_1^{surr}(f_1^*)}_{\mathtt{term~(d)}} + \underbrace{R_{ap}}_{\mathtt{term~(e)}}.
  \end{aligned}
  \label{eq:key_decomposition}
\end{equation}

The decomposition categorizes the error according to the sources they are incurred: $\mathtt{term~(a)}$ and $\mathtt{term~(d)}$ are the \emph{generalization error} due to the inaccessibility of the true data distribution, and $\mathtt{term~(b)}$ is the \emph{exploratory regret}, which not only includes the generalization error due to the limited budget to query the candidate features of the entire training dataset, but also includes the \emph{optimization error} due to the unknown best candidate feature in advance. The $\mathtt{term~(b)}$ of exploratory regret thus reflects the main difference between ExML and supervised learning. Besides, $\mathtt{term~(c)}$ is a negative term, and $\mathtt{term~(e)}$ is the unavoidable approximation error. This key decomposition shows that ExML not only requires to control the generalization error as SL does, but also needs to have a low exploratory regret, which has not been considered in previous study. In the remaining of this section, we will show the power of our feature exploration algorithms in lemmas, and verify the effectiveness of our proposed ExML approach in theorems. 

\subsubsection{Exploratory Learning with Uniform Allocation}
\label{sec:ExML-theory-ua}

According to the assumption on most informative feature which is introduced at the beginning of Section~\ref{sec:ExML-theory}, we succeed in identifying the best feature $a_1$ as long as we succeed to identify $a_1$ as the best feature in the entire training dataset. For ExML with feature exploration by uniform allocation (see details in Section~\ref{sec:feature-explore}), we have the following lemma that bounds the exploratory regret as shown in $\mathtt{term~(b)}$ of Eq.~\eqref{eq:key_decomposition}, 

\begin{myLemma}[Exploratory regret of uniform allocation]
  \label{lemma:ua_exploratory_regret}
  Let $a_{i_s}$ be the feature identified by uniform allocation, then uniform allocation identifies the best feature (i.e., $i_s=1$) with probability at least $1-\delta_{\text{fail}}$, where
    \begin{equation}
      \label{eq:failing-prob-UA}
      \delta_{\text{fail}} = 4(K-1)\exp\left(-\frac{2}{9}\lfloor B/K \rfloor\left(\frac{\Delta}{2} - \frac{2-2\theta}{1-2\theta}\sqrt{\frac{(\kappa\Lambda)^2}{\lfloor B/K \rfloor}}\right)^2\right),
    \end{equation}
    providing that the identification condition $\lfloor B/K \rfloor > \frac{16((1-\theta)\kappa\Lambda)^2}{((1-2\theta)\Delta)^2}$ holds, with $\theta$ the threshold of rejection model defined in~\eqref{eq:surrogate-loss}, $\Delta=\min_{i\in[K], \Delta_i>0}\Delta_i$ is the optimality gap defined in~\eqref{eq:optimality_gap}, $\Lambda=\sup_{i\in[K]}\Lambda_i$~and~$\kappa=\sup_{i\in[K]}\kappa_i$. 
    \newline Further more, for any $\delta>0$, with probability at least $1-\delta-\delta_{\text{fail}}$, we have
    $$\hat{R}_{tr, i_s}^{surr}(\hat{f}_{i_s})-\hat{R}_{tr, 1}^{surr}(\hat{f}_1^*)\leq \frac{4-4\theta}{1-2\theta}\sqrt{\frac{(\kappa\Lambda)^2}{\lfloor B/K \rfloor}}+2\sqrt{\frac{\log(2/\delta)}{2\lfloor B/K \rfloor}}~.$$
\end{myLemma}

\begin{myRemark}[Launch budget in feature exploration]
  Lemma~\ref{lemma:ua_exploratory_regret} bounds the exploratory regret induced by uniform allocation with high probability. We would notice that the identification condition introduces a ``launch budget'' for uniform allocation to be theoretically effective, and there is an extra probability $\delta_{\text{fail}}$ that uniform allocation would fail. These come from the statistical limit to differentiate features of different qualities with finite samples, and this statistical limit finally results in the difference between the excess risk bounds of ExML and supervised learning. \endenv
\end{myRemark}

Lemma~\ref{lemma:ua_exploratory_regret} directly yields a bound on $\mathtt{term~(b)}$ of Eq.~\eqref{eq:key_decomposition}, thus we can achieve the following theorem that validates the effectiveness of ExML equipped with uniform allocation: 

\begin{myThm}[Excess risk of ExML with uniform allocation]
\label{thm:uniform-allocation}
Let $a_{i_s}$ be the identified feature and $\hat{f}_{i_s}$ be the augmented model returned by \textsc{ExML} with uniform allocation. Then, for any $\delta > 0$, with probability at least $1-\delta-\delta_{\text{fail}}$, we have the following excess risk bound:
\begin{equation}
\label{eq:uniform-allocation-excess-risk}
R_{i_s}(\hat{f}_{i_s})-R_1^*\leq\O\left( \sqrt{\frac{(\kappa\Lambda)^2}{\lfloor B/K \rfloor}} + \sqrt{\frac{\log(6/\delta)}{2\lfloor B/K \rfloor}} \right) + R_{ap},
\end{equation}
with the failure probability $\delta_{\text{fail}} = \O\left( \exp\left(-\lfloor B/K \rfloor\right) \right)$ that decays exponentially with respect to the total budget $B$ (the formal definition can be found in~\eqref{eq:failing-prob-UA} of Lemma~\ref{lemma:ua_exploratory_regret}), providing that the identification condition $\lfloor B/K \rfloor > \frac{64((1-\theta)\kappa\Lambda)^2}{((1-2\theta)\Delta)^2}$ holds, where $\theta$ is the threshold of rejection model defined in~\eqref{eq:surrogate-loss}, $\Lambda=\sup_{i\in[K]}\Lambda_i$, $\kappa = \sup_{i\in[K]}\kappa_i$, and $R_{ap}$ is the approximation error introduced in~\eqref{eq:sl-excess-risk}.
\end{myThm}

\begin{myRemark}[Comparison between excess risk of SL and ExML]
  \label{remark:thm1-comparison}
We have the following comparison between the theoretical results of SL and ExML. Comparing the excess risk bounds of~\eqref{eq:sl-excess-risk} and~\eqref{eq:uniform-allocation-excess-risk}, we can observe that ExML exhibits a similar convergence tendency to SL with \emph{known} best feature yet \emph{without} requiring to know the best feature in advance, which is realized at the expense of an extra $\sqrt{K}$ times factor for the best feature exploration as well as an extra failure probability $\delta_{\text{fail}}$. 
\endenv 
\end{myRemark}

\subsubsection{Exploratory Learning with Median Elimination}

\label{sec:ExML-theory-me}

For ExML with feature exploration by median elimination~(Algorithm~\ref{alg:median-elimination} in Section~\ref{sec:feature-explore}), we have the following lemma that bounds the exploratory regret as shown in $\mathtt{term~(b)}$ of Eq.~\eqref{eq:key_decomposition},

\begin{myLemma}[Exploratory regret of median elimination]
  \label{lemma:me_exploratory_regret}
  Let $a_{i_s}$ be the feature identified by median elimination, then median elimination identifies the best feature (i.e., $i_s=1$) with probability at least $1-\delta_{\text{fail}}$, where
    \begin{equation}
      \label{eq:failing-prob-ME}
      \delta_{\text{fail}} =\frac{8\exp\left(-\frac{2}{9} \lfloor B/(K\log_2 K) \rfloor \left(\frac{\Delta}{2} - \frac{2-2\theta}{1-2\theta}\sqrt{\frac{(\kappa\Lambda)^2}{\lfloor B/(K\log_2 K) \rfloor}}\right)^2\right)}{1-\exp\left(-\frac{2}{9} \lfloor B/(K\log_2 K) \rfloor \left(\frac{\Delta}{2} - \frac{2-2\theta}{1-2\theta}\sqrt{\frac{(\kappa\Lambda)^2}{\lfloor B/(K\log_2 K) \rfloor}}\right)^2\right)}, 
    \end{equation}
    providing that the identification condition $\lfloor B/(K\log_2 K) \rfloor > \frac{16((1-\theta)\kappa\Lambda)^2}{((1-2\theta)\Delta)^2}$ holds, with $\theta$ the threshold of rejection model defined in~\eqref{eq:surrogate-loss}, $\Delta=\min_{i\in[K], \Delta_i>0}\Delta_i$ is the optimality gap defined in~\eqref{eq:optimality_gap}, $\Lambda=\sup_{i\in[K]}\Lambda_i$ ~and~ $\kappa=\sup_{i\in[K]}\kappa_i$. 
    \newline Further more, with probability at least $1-\delta-\delta_{\text{fail}}$, we have
    $$\hat{R}_{tr, i_s}^{surr}(\hat{f}_{i_s})-\hat{R}_{tr, 1}^{surr}(\hat{f}_1^*)\leq \frac{4-4\theta}{1-2\theta}\sqrt{\frac{(\kappa\Lambda)^2}{\lfloor B/\log_2 K \rfloor}}+2\sqrt{\frac{\log(2/\delta)}{2\lfloor B/\log_2 K \rfloor}}~.$$
\end{myLemma}

Lemma~\ref{lemma:me_exploratory_regret} directly yields a bound on $\mathtt{term~(b)}$ of Eq.~\eqref{eq:key_decomposition}, thus we can achieve the following theorem that validates the effectiveness of ExML equipped with median elimination~(Algorithm~\ref{alg:median-elimination}): 

\begin{myThm}[Excess risk of ExML with median elimination]
  \label{thm:median-elimination}
  Let $a_{i_s}$ be the identified feature and $\hat{f}_{i_s}$ be the augmented model returned by \textsc{ExML} with median elimination. Then, for any $\delta>0$, with probability at least $1-\delta-\delta_{\text{fail}}$, we have the following excess risk bound:
  \begin{equation}
  \label{eq:median-elimination-excess-risk}
  R_{i_s}(\hat{f}_{i_s})-R_1^*\leq\O\left( \sqrt{\frac{(\kappa\Lambda)^2}{\lfloor B/(\log_2 K) \rfloor}} + \sqrt{\frac{\log(6/\delta)}{2\lfloor B/(\log_2 K) \rfloor}} \right) + R_{ap},
  \end{equation}
  with the failure probability $\delta_{\text{fail}} = \O\left( \exp\left(-\lfloor B/(K\log_2 K) \rfloor\right) \right)$ which decays exponentially with respect to the total budget $B$ (the formal definition can be found in~\eqref{eq:failing-prob-ME} in Lemma~\ref{lemma:me_exploratory_regret}), providing that the identification condition $\lfloor B/(K\log_2 K) \rfloor > \frac{64((1-\theta)\kappa\Lambda)^2}{((1-2\theta)\Delta)^2}$ holds, where $\theta$ is the threshold of rejection model defined in~\eqref{eq:surrogate-loss}, $\Lambda=\sup_{i\in[K]}\Lambda_i$, $\kappa = \sup_{i\in[K]} \kappa_i$, and $R_{ap}$ is the approximation error in~\eqref{eq:sl-excess-risk}.
\end{myThm}

The proof of Theorem~\ref{thm:median-elimination} mostly parallels with that of Theorem~\ref{thm:uniform-allocation}, which includes a decomposition of excess risk as shown in Section~\ref{sec:ExML-decomposition} and a key lemma that bounds the exploratory regret induced by median elimination as shown in Lemma~\ref{lemma:me_exploratory_regret}. 

\begin{myRemark}[Comparison between uniform allocation and median elimination]
Comparing Theorem~\ref{thm:uniform-allocation} and Theorem~\ref{thm:median-elimination}, we can see that median elimination improves the $\sqrt{K}$ times factor paid for the feature exploration by uniform allocation to $\sqrt{\log_2 K}$ in the excess risk bound, as poor candidate features have been removed in the earlier episodes. By contrast, median elimination requires a larger ``launch budget'' in the identification condition compared to uniform allocation, and have a higher failure probability $\delta_{fail}$, because only a partial budget is used in early stages and so the best feature has a larger probability to be mistakenly discarded in the earlier episodes. Nevertheless, the failure probability in both results decays exponentially with respect to the total budget, which is thus low-order term and can be ignored in many situations. 
\endenv
\end{myRemark}

\section{Related Work}
\label{sec:related-work}
In this section, we briefly discuss some topics related to our proposed ExML framework.

\paragraph{Open Category Learning} Open category learning is also named as learning with new classes, which focuses on handling unknown classes appearing only in the testing phase~\citep{journals/pami/ScheirerRSB13,journals/pami/ScheirerJB14,conf/aaai/DaYZ14,conf/icml/LiuGDFH18,arXiv'19:LAC}, see the recent survey~\cite{arXiv'18:open-set-Huang} for a thorough overview of literature. Although these studies also care about the unknown classes detection, they differ from us significantly and thus cannot apply to our more challenging scenario: on one hand, they do not consider the issue of feature deficiency in the training data, which leads to great challenge in our problem; on the other hand, there exist unknown classes in the training data in our setting, while for open category learning the unknown classes only appear in the testing stage.

\paragraph{Learning with Unknown Unknowns} How to deal with unknown unknowns is a fundamental problem of robust artificial intelligence~\citep{TGD:robust-AI} and open-environment machine learning~\citep{nsr'22:Open-Survey,AAAI'20:open-world-learning}. A line of works deal with \emph{high-confidence false predictions} appear due to model's unawareness of such kind of mistake, which are also referred to as a kind of ``unknown unknowns''~\citep{journals/jqid/AttenbergPF15,conf/aaai/LakkarajuKCH17,conf/aaai/BansalW18}. Existing studies typically ask for external human expert to help identifying high-confidence false predictions and then retrain the model with the guidance.  Although these works also consider unknown unknowns and resort to external human knowledge, their setting and methodology differ from ours: our unknown unknowns are caused due to feature deficiency, so the learner requires to augment features rather than querying labels. Another kind of related works consider to \emph{avoid negative side effects}, which means that the reward functions in the prediction/decision process may be misleading due to the incomplete knowledge of the environments. There are emerging works that aim to detect and avoid the problem of negative side effects~\citep{IJAGI'12:NSE,NIPS'17:NSE,IJCAI'18:NSE,NIPS'20:NSE}. These works and ours both aim to enhance the robustness of AI systems in the face of unknown unknowns, while the specific problem modeling and developed methodologies are significantly different.

\paragraph{Active Learning} Active learning aims to achieve greater accuracy with fewer labels by asking queries of unlabeled data to be labeled by the human expert~\citep{book'12:active-learning}. Active learning bares certain similarities with our exploratory learning in the spirit --- instead of learning in a purely passive way, we both resort to some additional information sources to help the learning process. Interestingly, there are also some works querying features~\citep{conf/icdm/MelvillePM05,journals/ml/DhurandharS15,conf/kdd/HuangXXSNC18} to improve learning with missing features via as fewer as possible queries of entry values (feature of an instance). However, unlike their setting, we augment new features to help the identification of the unknown classes rather than querying missing values of the given feature to improve the performance of known classes classification. 

\paragraph{Learning with Rejection} Learning with rejection gives the classifier an option to reject an instance instead of providing a low-confidence prediction~\citep{journals/tit/Chow70}. Plenty of works are proposed to design effective algorithms~\citep{journals/jmlr/YuanW10,NIPS16:boosting-reject,conf/nips/WangQ18,conf/nips/ShimHY18,arxiv'22:active_learning} and establish theoretical foundations~\citep{journals/JASA/ZhangWX2018,Neurips'19:multiclass-rejection,ALT16:reject-theory,Stats06:reject-theory,journals/jmlr/BartlettW08,arXiv'19:bousquet-rejection}. As aforementioned, methods of learning with rejection cannot be directly applied in exploratory machine learning since it will result in inaccurate rejections of instances from known classes, and meanwhile, it cannot exploit new features. 

\paragraph{Feature Budget Learning} Feature budget learning considers a variant of supervised learning where an access of each feature on each sample is attached a cost, and the goal is to minimize the error within a given budget. This subject is initiated in~\citep{COLT'93:FBL}. \citet{ICML'12:FBL} pioneered the study of this area in linear regression considering uniform costs, and \citet{ICML'15:FBL} generalizes the results into the cases with non-uniform cost. The feature exploration module in our proposed approach is related to the setting in feature budget learning, while their results are restricted to specific choices of loss functions in order to get strong theoretical guarantees. Nevertheless, we believe it is possible to integrate the techniques of feature budget learning to develop more adaptive mechanisms in identifying top-$k$ features. 
\section{Experiments}
\label{sec:experiment}
In this section, we conduct experiments to examine empirical performance of the proposed exploratory machine learning (ExML). Specifically, we provide evaluations on synthetic data for visualizing the superiority of ExML to conventional supervised learning in handling unknown unknowns. Then, we report results on real-world datasets to demonstrate the effectiveness of the overall method, as well as the usefulness of feature exploration and model cascade modules.

The rejection models are learned with Gaussian kernel $K(\x_i,\x_j) = \exp(- \Vert\x_i-\x_j\Vert^2_2/\gamma )$, where the bandwidth $\gamma$ is set as $\gamma = \mbox{median}_{\x_i,\x_j\in D}(\Vert\mathbf{x}_i-\mathbf{x}_j\Vert_2^2)$. Besides, parameters $C_h$, $C_g$ are set as $1$. We select the best rejection threshold $\theta$ of augmented model from the pool $[0.1,0.2,0.3,0.4]$ for each algorithm, and threshold of the initial model is selected by cross validation to ensure 95\% accuracy on high-confidence predictions. Feature exploration budget is set as $B = b\cdot mK$, where $m$ is number of training samples, $K$ is number of candidate features, $b \in [0,1]$ is the budget ratio.

\begin{myRemark}[Automatic parameter tuning]
  We repeated the experiments by running ExML with each parameter settings in the pool, and the reported performance of each algorithm is the performance under their individual optimal parameters in hindsight. In fact, we can also perform automatic parameter tuning on the augmented model. For example, we can firstly use the best parameters of initial model to spend a proportion of budget on feature exploration to build a validation dataset, then select the best parameters by cross-validation on this dataset. \endenv
\end{myRemark}

\subsection{Synthetic Data for Illustration}
\label{sec:exp-synthetic}
We first illustrate the advantage of exploratory machine learning over the conventional supervised learning in discovery of the hidden classes on the synthetic data.

\begin{figure}[!t]
\centering
   \begin{subfigure}[ground-truth]{
    \begin{minipage}[b]{0.31\textwidth}
    \label{fig:truth}
     \includegraphics[clip, trim=3.4cm 10.2cm 4.0cm 9.2cm,width=\textwidth]{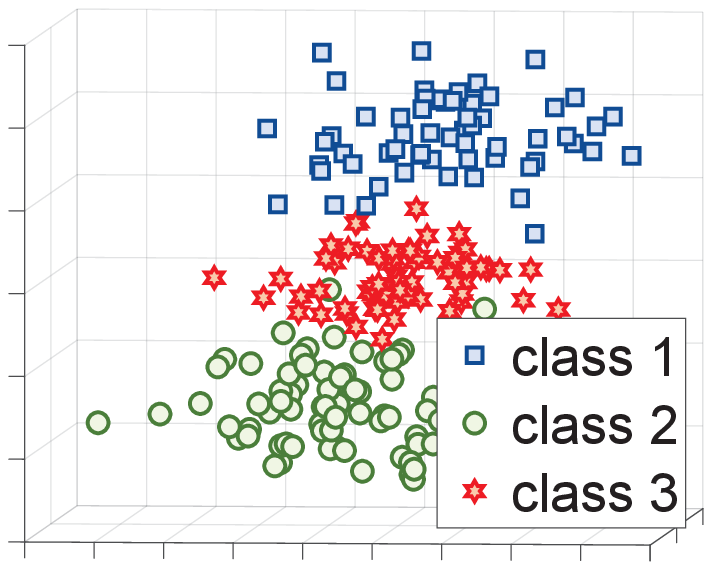}
     \end{minipage}}
  \end{subfigure}
  \begin{subfigure}[training data]{
    \begin{minipage}[b]{0.31\textwidth}
    \label{fig:syn-train}
     \includegraphics[clip, trim=3.4cm 10.2cm 4.0cm 9.2cm,width=\textwidth]{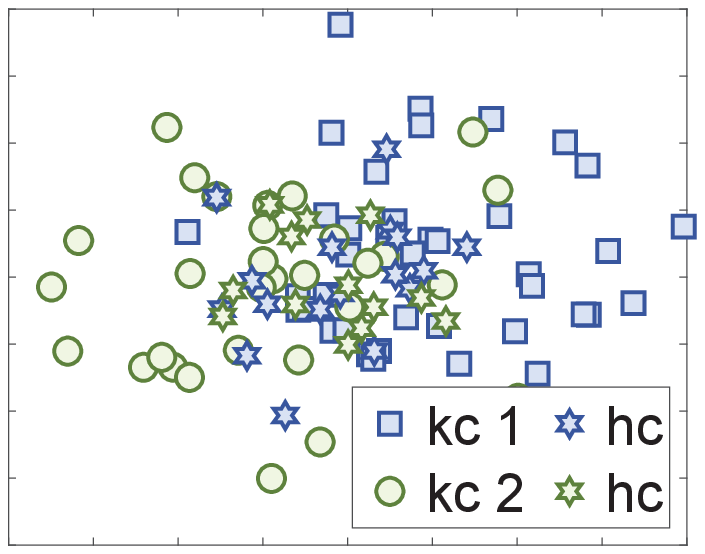}
     \end{minipage}}
  \end{subfigure}
  \begin{subfigure}[candidates]{
    \begin{minipage}[b]{0.31\textwidth}
    \label{fig:candidate-feature-pool}
     \includegraphics[clip, trim=3.4cm 10.2cm 4.0cm 9.2cm,width=\textwidth]{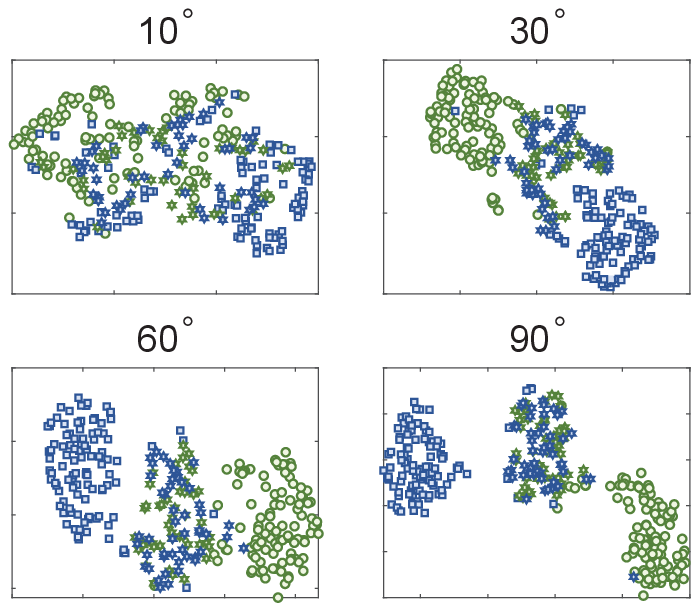}
     \end{minipage}}
  \end{subfigure}  
  \caption{Visualization of synthetic data. (a): ground-truth distribution; (b): training data (only first two dims are observable); (c): $t$-SNE of candidate features with various qualities (larger angles imply better features).}
\end{figure}

\paragraph{Setting} Following the illustrative example in Figure~\ref{fig:example}, we generate data with $3$-dim feature and $3$ classes, each class has $100$ samples. Figure~\ref{fig:truth} presents the ground-truth distribution. However, as shown in Figures~\ref{fig:syn-train}, the third-dim feature is unobservable in training data, resulting in a hidden class (\textsf{hc}) located in the intersection area of known classes (\textsf{kc1} and \textsf{kc2}). Samples from \textsf{hc} are mislabeled as \textsf{kc1} or \textsf{kc2} randomly. In detail, instances from each class are generated from a 3-dim Gaussian distributions. The means and variances are $[-a,0,-z]$ and $\sigma\cdot\mathbf{I}_{3\times 3}$ for class 1, $[a,0,z]$ and $\sigma\cdot\mathbf{I}_{3\times 3}$ for class 2 as well as $[0,0,0]$ and $\sigma/2\cdot\mathbf{I}_{3\times 3}$ for class 3, where $\mathbf{I}_{3\times 3}$ is a $3\times 3$ identity matrix. We fix $\sigma = 3a$ and set $z = 5a$. In the training stage, the third-dim is unobservable and the third class is randomly labeled as another two. There are 100 instances for each class in the training data. 

Besides, we generate $9$ candidate features in various qualities, whose angle to the  horizon varies from $10^\circ$ to $90^\circ$, the larger the better. Figure~\ref{fig:candidate-feature-pool} plots the augmented feature space via $t$-SNE. The budget ratio is $b = 20\%$. In the testing stage, the learner requires to predict on the 3-dim data, where the third dimension is the selected candidate features.

\begin{figure}[!t]
\centering
   \begin{subfigure}[\textsc{SL}]{
    \begin{minipage}[c]{0.31\textwidth}
    \label{fig:synthetic-SL}
     \includegraphics[clip, trim=3.4cm 9.4cm 4.0cm 9.8cm,width=\textwidth]{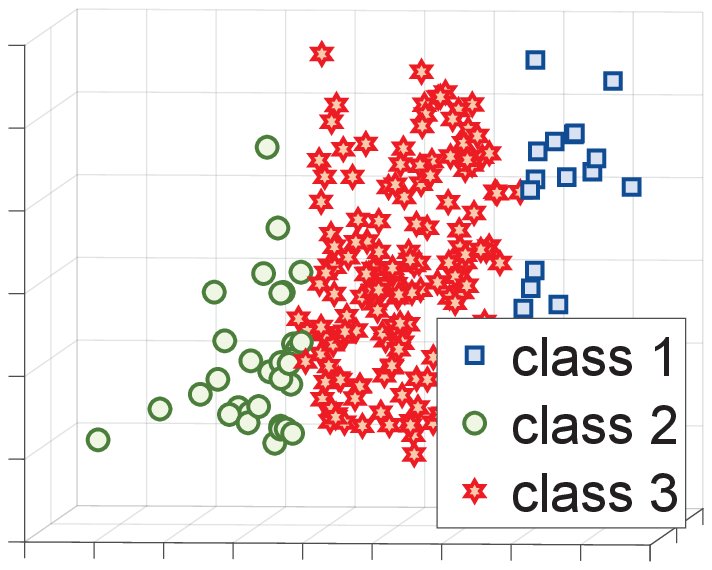} 
     \end{minipage}}
  \end{subfigure}
  \begin{subfigure}[\textsc{ExML}]{
    \begin{minipage}[d]{0.31\textwidth}
    \label{fig:our}
     \includegraphics[clip, trim=3.4cm 9.4cm 4.0cm 9.8cm,width=\textwidth]{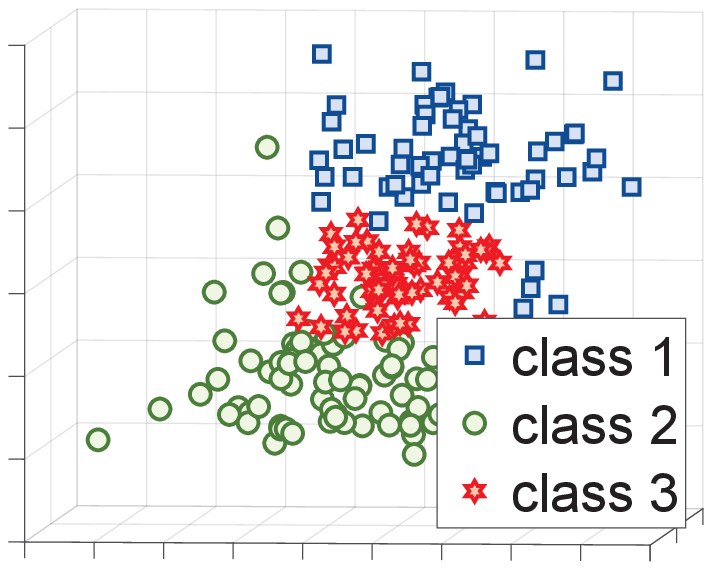}
     \end{minipage}}
  \end{subfigure}
  \begin{subfigure}[allocation]{
    \begin{minipage}[d]{0.31\textwidth}
    \label{fig:syn-allocation}
     \includegraphics[clip, trim=3.4cm 9.4cm 4.0cm 9.8cm,width=\textwidth]{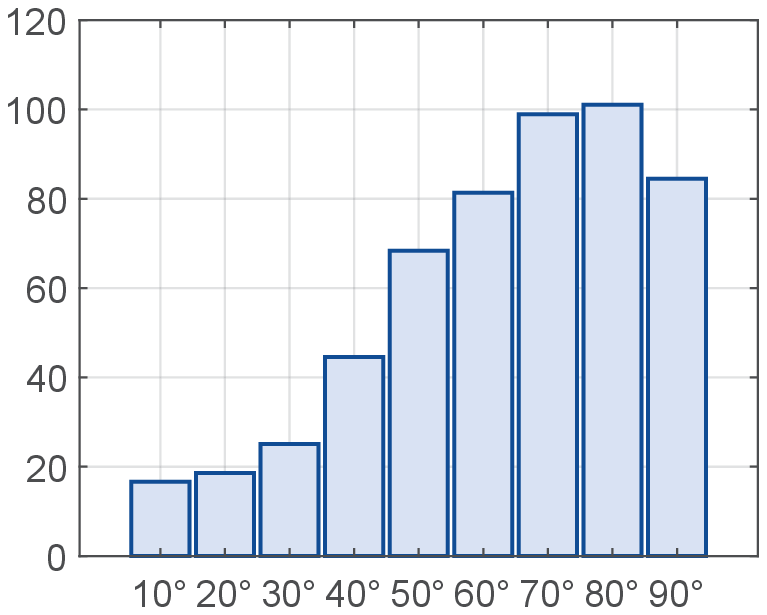}
     \end{minipage}}
  \end{subfigure}
  \caption{Visualization of results. (a)/(b): SL/ExML; (c): budget allocation of ExML with median elimination.}
  \label{fig:syn-exp1}
\end{figure}

\paragraph{Contenders} There are two contenders for the synthetic experiments, namely SL and ExML. For all the rejection model used in the experiments, we employ the Gaussian kernel with the bandwidth $\gamma = \mbox{median}_{\x_i,\x_j\in D}(\Vert\mathbf{x}_i-\mathbf{x}_j\Vert_2^2)$, and parameters $C_h, C_g$ are set to 1. 
\begin{compactitem}
\item \textbf{SL}: the rejection model~\citep{ALT16:reject-theory} trained on the 2-dim labeled training data, following the paradigm of conventional supervised learning. The threshold $\theta$ is chosen as one achieving best accuracy on the testing data from the pool $[0.1,0.2,0.3,0.4]$.
\item \textbf{ExML}: our proposal with cascade models and using median elimination for feature exploration. The threshold for the initial rejection model is selected by cross validation  to ensure 95\% accuracy on high-confidence samples. The threshold for the augmented rejection model is chosen as one achieving best accuracy on the testing data from the pool $[0.1,0.2,0.3,0.4]$. The budget ratio is $20\%$.
\end{compactitem}

\paragraph{Results}  We first conduct \textsc{SL} to train a rejection model based on the $2$-dim training data, and then perform \textsc{ExML} to actively augment the feature within the budget to discover unknown unknowns. Figures~\ref{fig:synthetic-SL} and~\ref{fig:our} plot the results, demonstrating a substantial advantage of \textsc{ExML} over \textsc{SL} in discovering the hidden class and predicting known classes. Furthermore, Figure~\ref{fig:syn-allocation} reports budget allocation of each candidate feature over $50$ times repetition. We can see that the allocation clearly concentrates to more informative features (with larger angles), which validates the effectiveness of median elimination for the best feature exploration. 

\subsection{Benchmark Data for Evaluation}
\label{sec:exp-mfeat}

We further evaluate on a UCI benchmark dataset \emph{Mfeat}~\citep{journals/kybernetika/BreukelenDTH98}. 

\paragraph{Dataset} Mfeat is a multi-view dataset\footnote{The dataset can be downloaded from \url{http://archive.ics.uci.edu/ml/datasets/Multiple+Features}} containing 2000 samples and 6 views of features extracted by various methods, whose brief semantic and statistical information are listed as follows. 
\begin{compactitem}
  \item \textbf{Fac}: profile correlations, 216-dim;
  \item \textbf{Pix}: pixel averages in $2\times3$ windows, 240-dim;
  \item \textbf{Kar}: Karhunen-Love coefficients, 64-dim;
  \item \textbf{Zer}: Zernike moments, 47-dim;
  \item \textbf{Fou}: Fourier coefficients of the character shapes, 76-dim;
  \item \textbf{Mor}: morphological features, 6-dim.    
\end{compactitem}
According to the domain knowledge, we can sort the six features by their feature quality as: Fac $>$ Pix $>$ Kar $>$ Zer $>$ Fou $>$ Mor, in a descending order.

Since \emph{Mfeat} is a multi-class dataset, we randomly sample 5 configurations to convert it into the binary classification task, where each known class and hidden class contain three original classes (and so each configuration includes an amount of 1800 samples), and the instances from the hidden class are randomly mislabeled as one of known classes. There are in total 50 random configurations for training. As for the candidate features, we take one as original and the rest are prepared in the candidate set. Before training, we normalize all the features to the range $[0,1]$. In the training stage, 600 instances are randomly samples from the whole dataset for 10 times to form the labeled training data. In the testing stage, the rest 1200 instances are used for measuring the performance of compared algorithms. 

\paragraph{Setting}  We randomly sample 600 instances as the training data for 10 times, and the rest are used for testing. As for the candidate features, each one of six views (features) is taken as original feature and the rest are prepared as candidate features. The budget ratio varies from 10\% to 30\%. 

\paragraph{Contenders} Apart from SL, we include two ExML variants: \textsc{ExML}$_{\text{csd}}^{\text{UA}}$ and \textsc{ExML}$_{\text{aug}}^{\text{ME}}$ for ablation studies. Here aug/csd denotes the final model is only the augmented or  cascaded with the initial model; UA/ME refers to feature exploration by uniform allocation or median elimination. 
\begin{compactitem}
  \item \textbf{\textsc{ExML}$_{\text{csd}}^{\text{UA}}$}: our proposal with cascade model and using \emph{uniform allocation} for feature exploration, sharing the same parameters setting as ExML.
  \item \textbf{\textsc{ExML}$_{\text{aug}}^{\text{ME}}$}: our proposal \emph{without} cascade model and using median elimination for feature exploration, sharing the same parameters setting as ExML.
\end{compactitem}

For all ExML methods, the budget ratio $b$ varies from 10\% to 30\%. The parameter settings of SL and ExML are the same as those in the synthetic experiments (Section~\ref{sec:exp-synthetic}).

\paragraph{Measure} We measure the performance of all the methods by the classification. Additionally, we introduce the \emph{recall} to measure the effectiveness of feature exploration, defined as the ratio of the number of cases when identified feature is one of its top 2 features to the total number.
\begin{compactitem}
  \item \textbf{Accuracy}: the mean and standard deviation of the predictive accuracy on testing dataset over 50 configurations, where the true label of hidden classes are observable.
  \item \textbf{Recall}: the ratio of the number of cases when identified feature is one of its top 2 features to the total number, where the quality of features is measured by the attainable accuracy of the augmented model trained on the whole dataset with this particular feature.
\end{compactitem}

\begin{table}[!t]
\caption{Evaluation on \emph{Mfeat} dataset. Features are sorted by descending feature qualities. Bold font indicates algorithms significantly outperforms than others (paired $t$-test at 95\% significance level).}
\vspace{3mm}
\label{table:benchmark}
\scriptsize
\centering
\resizebox{0.99\textwidth}{!}{
\begin{tabular}{cccccccc}\toprule
Feature                   &Description                                                & Budget  & \textsc{SL} & \textsc{ExML}$_{\text{aug}}^{\text{ME}}$     & \textsc{ExML}$_{\text{csd}}^{\text{UA}}$     & \textsc{ExML} (= \textsc{ExML}$_{\text{csd}}^{\text{ME}}$)  & Recall \\ \midrule                                                 
\multirow{3}{*}{Fac}      &\multirow{3}{*}{Profile correlations}                      & 10\%    & \textbf{93.39} $\pm$ \textbf{1.66}         & 71.80 $\pm$ 9.55                             & 92.39 $\pm$ 2.79                             & 92.40 $\pm$ 2.78                       & 48\%    \\      
                          &                                                           & 20\%    & \textbf{93.39} $\pm$ \textbf{1.66}         & 82.26 $\pm$ 7.52                             & 91.95 $\pm$ 3.32                             & 92.00 $\pm$ 3.27                       & 46\%    \\             
                          &                                                           & 30\%    & \textbf{93.39} $\pm$ \textbf{1.66}         & 89.29 $\pm$ 4.72                             & 92.20 $\pm$ 3.33                             & 92.50 $\pm$ 2.86                       & 44\%    \\ \midrule           
\multirow{3}{*}{Pix}      &\multirow{3}{*}{ \begin{tabular}[c]{@{}c@{}}Pixel averages\\ in 2 $\times$ 3 windows\end{tabular}}           & 10\%    & \textbf{92.19} $\pm$ \textbf{2.47}         & 70.53 $\pm$ 8.27                             & 90.54 $\pm$ 6.27                             & 90.55 $\pm$ 6.31                       & 58\%    \\   
                          &                                                           & 20\%    & \textbf{92.19} $\pm$ \textbf{2.47}         & 81.70 $\pm$ 7.16                             & 90.84 $\pm$ 6.17                             & 90.87 $\pm$ 6.09              &  54\%  \\                        
                          &                                                           & 30\%    & \textbf{92.19} $\pm$ \textbf{2.47}         & 88.67 $\pm$ 4.14                             & 90.45 $\pm$ 5.74                             & \textbf{91.82} $\pm$ \textbf{4.26}     & 68\%  \\ \midrule  
\multirow{3}{*}{Kar}      &\multirow{3}{*}{ \begin{tabular}[c]{@{}c@{}}Karhunen-Love \\ coefficients\end{tabular}}            & 10\%    & \textbf{86.87} $\pm$ \textbf{3.43}         & 70.25 $\pm$ 10.2                            & 85.55 $\pm$ 4.94                             & 85.90 $\pm$ 4.85                        & 56\%   \\                                         
                          &                                                           & 20\%    & \textbf{86.87} $\pm$ \textbf{3.43}         & 81.46 $\pm$ 6.88                             & 85.21 $\pm$ 5.46                             & \textbf{86.49} $\pm$ \textbf{4.81}     & 54\%   \\                             
                          &                                                           & 30\%    & 86.87 $\pm$ 3.43                             & 86.01 $\pm$ 5.41                             & 86.52 $\pm$ 4.71                             & \textbf{88.18} $\pm$ \textbf{3.57}   & 56\%    \\ \midrule   
\multirow{3}{*}{Zer}      &\multirow{3}{*}{Zernike moments}                           & 10\%    & 73.82 $\pm$ 8.82                             & 69.61 $\pm$ 10.7                             & 72.96 $\pm$ 10.4                             & \textbf{76.17} $\pm$ \textbf{8.52}   & 82\%   \\                             
                          &                                                           & 20\%    & 73.82 $\pm$ 8.82                             & \textbf{80.86} $\pm$ \textbf{8.02}         & 77.31 $\pm$ 7.89                             & \textbf{81.72} $\pm$ \textbf{7.33}     & 82\%   \\                       
                          &                                                           & 30\%    & 73.82 $\pm$ 8.82                             & \textbf{86.07} $\pm$ \textbf{5.51}         & 81.11 $\pm$ 6.79                             & \textbf{86.33} $\pm$ \textbf{5.04}     & 86\%   \\ \midrule   
\multirow{3}{*}{Fou}      &\multirow{3}{*}{Fourier coefficients}                      & 10\%    & 68.73 $\pm$ 9.07                             & 69.42 $\pm$ 9.68                             & 68.88 $\pm$ 11.8                             & \textbf{75.92} $\pm$ \textbf{8.81}   & 82\%   \\                       
                          &                                                           & 20\%    & 68.73 $\pm$ 9.07                             & 82.11 $\pm$ 6.48                             & 77.93 $\pm$ 8.27                             & \textbf{85.03} $\pm$ \textbf{4.39}   & 88\%   \\                       
                          &                                                           & 30\%    & 68.73 $\pm$ 9.07                             & \textbf{89.90} $\pm$ \textbf{3.69}          & 82.45 $\pm$ 5.20                             & \textbf{89.35} $\pm$ \textbf{3.89}    & 92\%   \\ \midrule  
\multirow{3}{*}{Mor}      &\multirow{3}{*}{Morphological features}                    & 10\%    & 57.47 $\pm$ 15.3                             & 69.09 $\pm$ 11.3                             & 66.58 $\pm$ 13.5                             & \textbf{71.07} $\pm$ \textbf{11.1}   & 80\%    \\                        
                          &                                                           & 20\%    & 57.47 $\pm$ 15.3                             & \textbf{79.60} $\pm$ \textbf{10.1}         & 73.61 $\pm$ 8.86                             & \textbf{79.74} $\pm$ \textbf{9.92}     & 84\%   \\                               
                          &                                                           & 30\%    & 57.47 $\pm$ 15.2                             & \textbf{87.44} $\pm$ \textbf{7.34}         & 78.31 $\pm$ 9.00                             & \textbf{86.98} $\pm$ \textbf{7.07}     & 90\%   \\ \bottomrule 
  \vspace{-4mm}
\end{tabular}}  
\end{table}

\paragraph{Results} Table~\ref{table:benchmark} reports mean and std of the predictive accuracy, and all features are sorted in descending order by their quality. We first compare the conventional supervised learning (SL) to (variants of) ExML. When the original features are in high quality (Kar, Pix, Fac), SL could achieve favorable performance and there is no need to explore new features. However, in the case where uninformative original features are provided, which is of more interest for ExML, SL degenerates severely and \textsc{ExML}$_{\text{aug}}^{\text{ME}}$ (the single ExML model without model cascade) achieves better performance even with the limited budget. Besides, from the last column, we can see that informative candidates (top 2) are selected to strengthen the poor original features, which validates the efficacy of the proposed budget allocation strategy.

Since the \textsc{ExML}$_{\text{aug}}^{\text{ME}}$ is not guaranteed to outperform SL, particularly with the limited budget on poor candidate features, we propose the cascade structure. Actually, ExML approach (aka, \textsc{ExML}$_{\text{csd}}^{\text{ME}}$) achieves roughly \emph{best-of-two-worlds} performance, in the sense that it is basically no worse or even better than the best of SL and \textsc{ExML}$_{\text{aug}}^{\text{ME}}$. It turns out that even \textsc{ExML}$_{\text{csd}}^{\text{UA}}$ could behave better than \textsc{ExML}$_{\text{aug}}^{\text{ME}}$. These results validate the effectiveness of the model cascade component. 

Notice that there are also some cases that the augmented model (\textsc{ExML}$_{\text{aug}}^{\text{ME}}$) outperforms the cascade model (\textsc{ExML}$_{\text{aug}}^{\text{ME}}$). Indeed, since the rejection model at the first layer is trained on the original dataset, the performance of the cascaded model will be affected by the rejection model to some extent. When feature exploration is of high quality, the performance of the second layer itself already becomes good enough, the model cascade may slightly affect the overall performance. Nevertheless, the cascading structure can still prevent the impact of low-quality feature exploration on the overall performance, enhancing the robustness of our method.

\subsection{Real Data of Activities Recognition}
\label{sec:exp-real-data}
We additionally examine the effectiveness of our proposed algorithm on a real-world dataset called \emph{RealDisp}\footnote{\url{http://archive.ics.uci.edu/ml/datasets/REALDISP+Activity+Recognition+Dataset}}, which is an activities recognition task~\citep{conf/huc/BanosDPRTA12}. Specifically, there are 9 on-body sensors used to capture various actions of participants. Each sensor is placed on different parts of the body and provides 13-dimensional features including 3-dim from acceleration, 3-dim from gyro, 3-dim from  magnetic field orientation and another 4-dim from quaternions. Hence, in this dataset we have 117 features in total.

\paragraph{Dataset} In our experiments, three types of actions (\emph{walking}, \emph{running}, and \emph{jogging}) of the first subject under the ideal-placement are included to form the dataset containing 2000 instances, where 30\% of them are used for training and 70\% for testing. In the training data, one sensor is deployed and the class of jogging is mispercevied as walking or running randomly. The learner would explore the rest eight candidate features to discover the unknown unknowns. Thus, there are 9 partitions, and each is repeated for 10 times by sampling the training instances randomly.

\begin{figure}[!t]
\centering
 \begin{minipage}[t]{0.47\textwidth}
    \centering
    \includegraphics[clip, trim=0cm 0.2cm 0cm 0cm,height=0.76\textwidth]{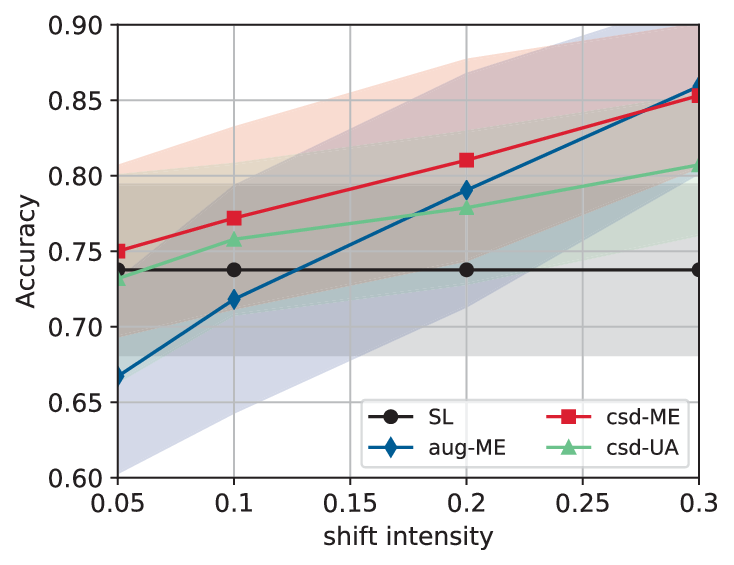}
    \setcaptionwidth{6cm}
    \vspace{-6.5mm}
    \caption{Performance comparisons of all the contenders.}
    \label{fig:realdisp_acc}
\end{minipage} 
 \begin{minipage}[t]{0.47\textwidth}
    \centering
    \includegraphics[clip, trim=4.3cm 9.0cm 4.5cm 9.8cm,height=0.75\textwidth]{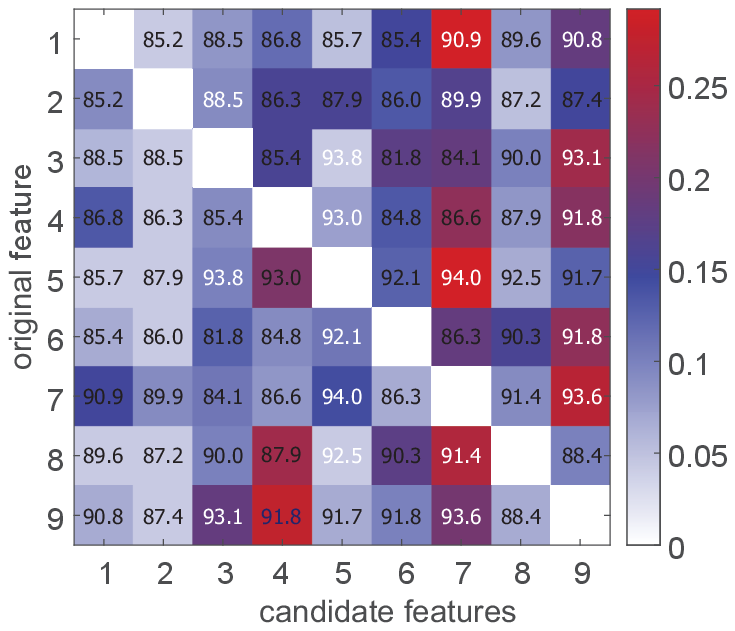}
    \setcaptionwidth{6cm}
    \caption{Illustration of budget allocation with median elimination.}
    \label{fig:sample_pool}
\end{minipage}  
\end{figure}

\paragraph{Results} Figure~\ref{fig:realdisp_acc} shows the mean and std of accuracy, our approach \textsc{ExML} (aka, \textsc{ExML}$_{\text{csd}}^{\text{ME}}$) outperforms others, validating the efficacy of our proposal. In addition, Figure~\ref{fig:sample_pool} illustrates the budget allocation when the budget ratio $b=30\%$. The $i$-th row denotes the scenario when the $i$-th sensor is the original feature, and patches with colors indicate the fraction of budget allocated to each candidate feature. The number above a patch means the attainable accuracy of the model trained on the whole training dataset with the particular feature. We highlight the top two candidate features of each row in white, and use blue color to indicate selected feature is not in top two. The results show that \textsc{ExML} with median elimination can select the top two informative features to augment for all the original sensors. The only exception is the 9-th sensor, but quality of the selected feature (91.8$\%$) does not deviate too much from the best one (93.6$\%$). These results reflect the effectiveness of our feature exploration strategy.

\subsection{Non-Uniform Cost}
We finally examine the effectiveness of our proposed principle for non-uniform cost on the previous \emph{Mfeat} dataset with each group of features attached with a different cost. 

\paragraph{Dataset} \emph{Mfeat} is a multi-view dataset containing 2000 samples and 6 views of features extracted by various methods, whose brief semantic and statistical information are listed as follows. Additionally, we attach each feature a cost which are shown in the brackets below, in order to simulate the non-uniform cost scenario. 
\begin{compactitem}
  \item \textbf{Fac} (5.0) : profile correlations, 216-dim;
  \item \textbf{Pix} (1.2) : pixel averages in $2\times3$ windows, 240-dim;
  \item \textbf{Kar} (1.0) : Karhunen-Love coefficients, 64-dim;
  \item \textbf{Zer} (0.95): Zernike moments, 47-dim;
  \item \textbf{Fou} (1.5) : Fourier coefficients of the character shapes, 76-dim;
  \item \textbf{Mor} (0.9) : morphological features, 6-dim. 
\end{compactitem}
According to the domain knowledge, we can sort the six features by their feature quality as: Fac $>$ Pix $>$ Kar $>$ Zer $>$ Fou $>$ Mor, in a descending order. 

We remark that our cost attachment is rational, as it includes a feature with highest quality but is expensive (Fac), features with relatively high quality and are cheap (Pix, Kar), features that are cheap but with low quality (Zer, Mor), and a feature with low quality and is expensive (Fou). Intuitively, the expected behavior of the algorithm is to query more samples on the relatively good and cheap features (Pix, Kar), rather than to spend a lot on expensive features which leads to poor generalization ability, nor to query a lot on inherently poor features. 

\paragraph{Setting}  Same as the experimental setup in Section~\ref{sec:exp-mfeat}, we randomly sample 600 instances as the training data for 10 times, and the rest are used for testing. As for the candidate features, each one of six views (features) is taken as original feature and the rest are prepared as candidate features. The budget ratio varies from 10\% to 30\%. 

\paragraph{Contenders} We include four ExML variants: \textsc{ExML}$_{\text{SA}}^{\text{UA}}$, \textsc{ExML}$_{\text{BA}}^{\text{UA}}$, \textsc{ExML}$_{\text{SA}}^{\text{ME}}$ and \textsc{ExML}$_{\text{BA}}^{\text{ME}}$ for ablation studies. Here SA/BA denotes the principle of non-uniform cost adaptation, where SA means sample alignment and BA means budget alignment; UA/ME refers to feature exploration by uniform allocation or median elimination. For all ExML methods, the parameter settings are the same as those in the synthetic experiments (Section~\ref{sec:exp-synthetic}). 

\begin{figure}[!t]
  \centering
  \begin{subfigure}[Sample allocation with original feature Fou]{
   \begin{minipage}[t]{0.85\linewidth}
      \centering
      \includegraphics[width=\linewidth]{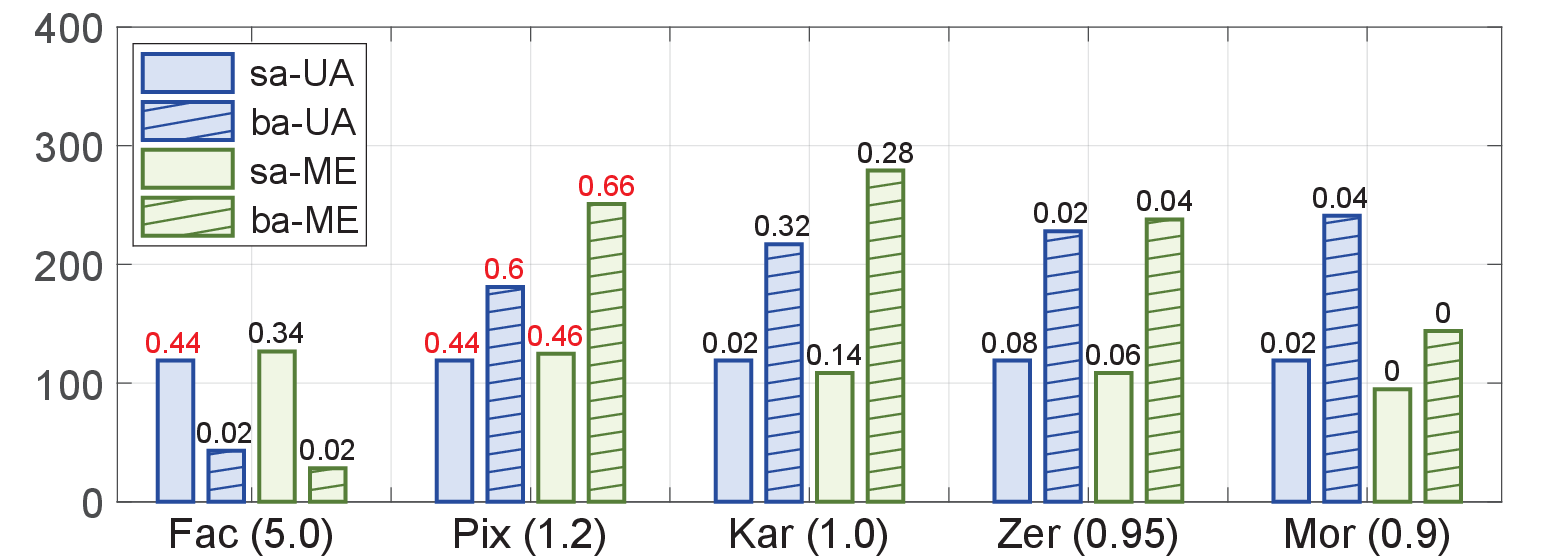}
      \label{fig:sample_allocation_Fou}
  \end{minipage}}
\end{subfigure}
  \begin{subfigure}[Sample allocation with original feature Zer]{
   \begin{minipage}[t]{0.85\linewidth}
      \centering
      \includegraphics[width=\linewidth]{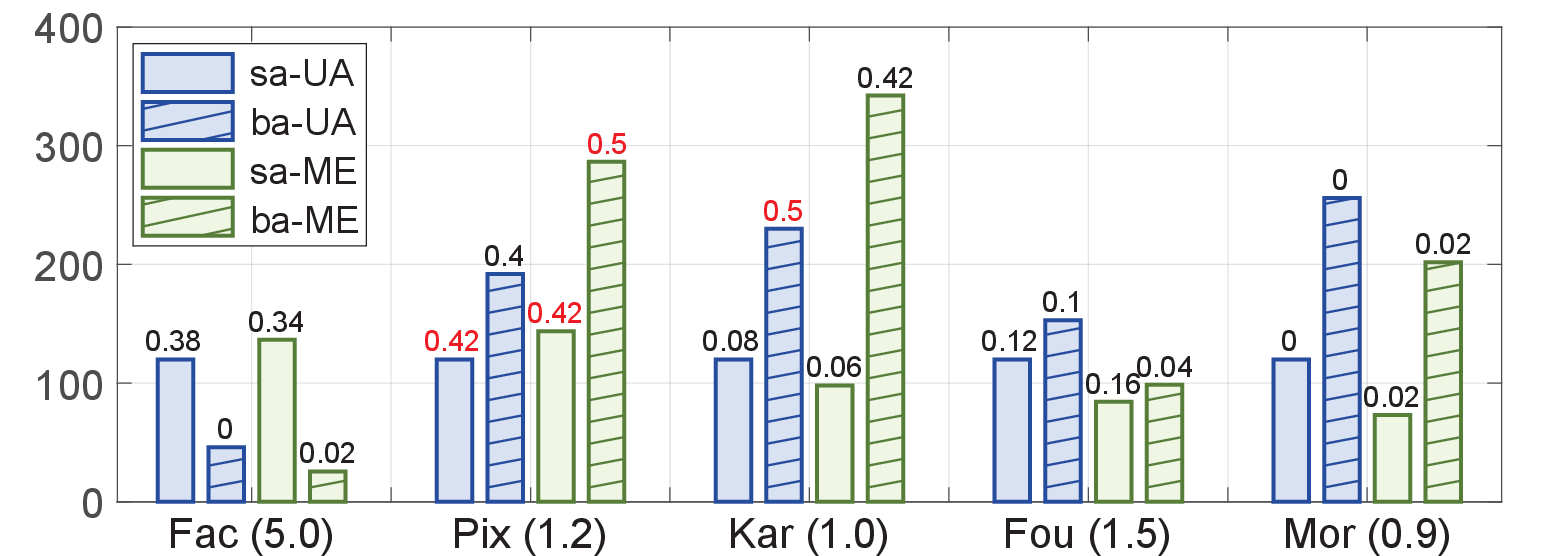}
      \label{fig:sample_allocation_Zer}
  \end{minipage}}
\end{subfigure}
  \vspace{-2mm}
  \caption{Sample allocation with different original features. The $x$-axis denotes the candidate features sorted by descending qualities and the values in brackets are their costs. The $y$-axis denotes the number of queries on each feature. The values over the bars are the ratios of the number of cases that the algorithm identifies the corresponding candidate as the best feature, with the reds indicating the most frequent ones. }
  \label{fig:sample_allocation}
\end{figure}

\begin{table}[!t]
  \caption{Evaluation on \emph{Mfeat} dataset attached with non-uniform costs. Features are sorted by descending feature qualities. Bold font indicates algorithms significantly outperforms than others (paired $t$-test at 95\% significance level).}
  \vspace{2mm}
  \label{table:non_uniform_benchmark}
  \scriptsize
  \centering
  \resizebox{0.99\textwidth}{!}{
  \begin{tabular}{ccccccc}\toprule
  Feature                   &Description                                                & Budget  & \textsc{ExML}$_{\text{SA}}^{\text{UA}}$ & \textsc{ExML}$_{\text{BA}}^{\text{UA}}$     & \textsc{ExML}$_{\text{SA}}^{\text{ME}}$     & \textsc{ExML}$_{\text{BA}}^{\text{ME}}$ \\ \midrule                                                 
  \multirow{3}{*}{Fac}      &\multirow{3}{*}{Profile correlations}                      & 10\%    & 91.81 $\pm$ 3.10         & 91.81 $\pm$ 3.10                             & 91.58 $\pm$ 5.74                             & 91.58 $\pm$ 5.74                           \\      
                            &                                                           & 20\%    & 92.30 $\pm$ 2.45         & 92.31 $\pm$ 2.44                             & 91.74 $\pm$ 3.32                             & 91.73 $\pm$ 3.25                           \\             
                            &                                                           & 30\%    & 92.13 $\pm$ 3.32         & 92.14 $\pm$ 3.32                             & 92.33 $\pm$ 2.82                             & 92.26 $\pm$ 2.83                           \\ \midrule           
  \multirow{3}{*}{Pix}      &\multirow{3}{*}{ \begin{tabular}[c]{@{}c@{}}Pixel averages\\ in 2 $\times$ 3 windows\end{tabular}}           & 10\%    & 90.90 $\pm$ 3.77         & 90.90 $\pm$ 3.77                             & 91.22 $\pm$ 3.61                             & 91.22 $\pm$ 3.62                          \\   
                            &                                                           & 20\%    & 90.33 $\pm$ 6.14         & 90.42 $\pm$ 5.27                             & 90.65 $\pm$ 4.52                             & 90.86 $\pm$ 4.28                \\                        
                            &                                                           & 30\%    & 90.96 $\pm$ 4.11         & 90.78 $\pm$ 4.23                             & 91.57 $\pm$ 3.06                             & \textbf{92.72} $\pm$ \textbf{2.51}       \\ \midrule  
  \multirow{3}{*}{Kar}      &\multirow{3}{*}{ \begin{tabular}[c]{@{}c@{}}Karhunen-Love \\ coefficients\end{tabular}}            & 10\%    & 84.27 $\pm$ 5.84         & 84.14 $\pm$ 6.09                            & 84.11 $\pm$ 5.80                             & 84.95 $\pm$ 6.02                           \\                                         
                            &                                                           & 20\%    & 84.64 $\pm$ 5.85         & 84.78 $\pm$ 6.55                             & 84.67 $\pm$ 5.62                             & \textbf{88.84} $\pm$ \textbf{3.74}       \\                             
                            &                                                           & 30\%    & 84.95 $\pm$ 5.36                             & 86.35 $\pm$ 4.96                             & 87.27 $\pm$ 4.09                             & \textbf{90.65} $\pm$ \textbf{3.08}       \\ \midrule   
  \multirow{3}{*}{Zer}      &\multirow{3}{*}{Zernike moments}                           & 10\%    & 70.99 $\pm$ 9.77                             & 70.72 $\pm$ 8.76                             & 74.16 $\pm$ 9.54                             & \textbf{79.42} $\pm$ \textbf{6.20}      \\                             
                            &                                                           & 20\%    & 76.02 $\pm$ 8.19                             & 76.92 $\pm$ 7.78         & 81.33 $\pm$ 7.98                             & \textbf{86.32} $\pm$ \textbf{6.77}        \\                       
                            &                                                           & 30\%    & 80.60 $\pm$ 6.88                             & 81.10 $\pm$ 6.60         & 85.93 $\pm$ 6.24                             & \textbf{90.59} $\pm$ \textbf{3.83}        \\ \midrule   
  \multirow{3}{*}{Fou}      &\multirow{3}{*}{Fourier coefficients}                      & 10\%    & 69.67 $\pm$ 9.52                             & 69.02 $\pm$ 10.3                             & 74.45 $\pm$ 7.72                             & \textbf{76.93} $\pm$ \textbf{7.65}      \\                       
                            &                                                           & 20\%    & 77.43 $\pm$ 6.96                             & 75.26 $\pm$ 7.22                             & 81.39 $\pm$ 7.48                             & \textbf{88.04} $\pm$ \textbf{3.12}      \\                       
                            &                                                           & 30\%    & 84.08 $\pm$ 4.30                             & 83.21 $\pm$ 5.13          & 87.57 $\pm$ 3.86                             & \textbf{90.17} $\pm$ \textbf{3.21}      \\ \midrule  
  \multirow{3}{*}{Mor}      &\multirow{3}{*}{Morphological features}                    & 10\%    & 63.61 $\pm$ 13.9                             & 65.59 $\pm$ 10.4                             & 68.95 $\pm$ 10.7                             & \textbf{74.31} $\pm$ \textbf{7.44}      \\                        
                            &                                                           & 20\%    & 73.04 $\pm$ 8.46                             & 72.85 $\pm$ 11.2         & 77.41 $\pm$ 9.90                             & \textbf{86.14} $\pm$ \textbf{6.72}        \\                               
                            &                                                           & 30\%    & 79.61 $\pm$ 8.84                             & 82.38 $\pm$ 7.44         & 85.32 $\pm$ 7.74                             & \textbf{90.28} $\pm$ \textbf{5.90}        \\ \bottomrule 
    \vspace{-4mm}
  \end{tabular}}
\end{table}

\paragraph{Results} Table~\ref{table:non_uniform_benchmark} reports mean and std of the predictive accuracy, and all features are sorted in descending order by their quality. As verified in previous experiments, when the original features are in high quality (Pix, Fac), all contenders mostly relies on the prediction of the first layer, which lead to similar performance. However, in the case where uninformative original features are provided, \textsc{ExML} achieves better performance with limited budget since features with better quality are explored, and within the four contenders, \textsc{ExML}$_{\text{BA}}^{\text{ME}}$ outperforms the other three algorithms because \textsc{ExML}$_{\text{BA}}^{\text{ME}}$ allocates more budget to relatively good but much cheaper features (Pix, Kar). 

Moreover, Figure~\ref{fig:sample_allocation} shows the sample allocation of each candidate feature over 50 random configurations with the budget ratio $b=20\%$. The colors of the bars indicate the basic budget allocation strategy, i.e., blue for uniform allocation and green for median elimination, and the shades in the bars indicate the non-uniform adaptation principles, i.e., the empty shade for sample allocation and the dot for budget allocation. Besides, the values over the bars are the ratios of the number of cases that the algorithm identifies the corresponding candidate as the best feature, with the red texts are the most frequent ones. We can see that the budget alignment principle (`ba-') avoids allocating too much budget to query expensive feature (Fac) comparing to the sample alignment principle (`sa-' with the empty shade). Besides, median elimination (`ME') shows a clearer concentration on relatively good but much cheaper features (Pix, Kar) comparing to uniform allocation (`UA'), which results in a better generalization ability. This verifies our interpretation of Table~\ref{table:non_uniform_benchmark} above, and validates the effectiveness of our median elimination strategy as well as the budget alignment principle proposed in Section~\ref{sec:feature-explore}. 
\section{Conclusion}
\label{sec:conclusion}
This paper studies the task of learning with unknown unknowns, where there exist some instances in training datasets belonging to an unknown hidden class but are wrongly perceived as known classes, due to the insufficient feature information. To address this issue, we propose the \emph{exploratory machine learning} (ExML) to encourage the learner to examine and investigate the training dataset by exploring more features to discover potentially hidden classes. Following this principle, we design an approach consisting of three procedures: rejection model, feature exploration, and model cascade. By leveraging techniques from bandit theory, we prove the rationale and efficacy of the feature exploration procedure. Experiments validate the effectiveness of our approach.

There remain many interesting directions to further push forward the study of exploratory machine learning. First, as mentioned, one may borrow more advanced techniques to relax some current modeling assumptions such as binary known classes, best feature exploration, etc. Second, the method proposed in this paper is merely one implementation of the ExML framework, and exploring other effective mechanisms of feature exploration and hierarchical processing is also left as an interesting future work. Third, since in the environments with unknown unknowns, it would be difficult to expect passive learning can do well and the algorithm should explore necessary additional information from the environments, we believe the methodology behind our proposed ExML framework can serve as a principled way to handle unknown unknowns even beyond the scope of our concerned one due to feature deficiency. 

Furthermore, unknown unknowns not only appear in the tasks of prediction, but also in the field of decision making. There are paradigms that models the sequential decision-making processes, such as reinforcement learning and rehearsal learning. In reinforcement learning, an agent learns to make decisions by performing actions in an environment to achieve maximum cumulative reward~\citep{book'18:RL}. As for rehearsal learning, the learner tries to act proactively to prevent undesirable outcomes, which is a promising domain for further exploration~\citep{FCS'22:rehearsal}. Evidently, the unknown unknowns issue becomes even more severe in decision-making tasks compared to the prediction tasks, because the effect of unknown unknowns at current decision stage may entangle with the effect of unknown unknowns in the past stages. We believe that the methodology behind our proposed ExML framework, especially the principle of interactively exploring more information from environments, can be extended to decision-making scenarios. 

\section*{Acknowledgment}
This research was supported by NSFC (U23A20382, 61921006), JiangsuSF (BK20220776), the Collaborative Innovation Center of Novel Software Technology and Industrialization, National Postdoctoral Program for Innovative Talent, and China Postdoctoral Science Foundation (2023M731597). We are grateful for the anonymous reviewers from AAAI and AIJ for their valuable comments.

\bibliography{explore_abbr}
\bibliographystyle{elsarticle-num-names}

\clearpage
\appendix
\section*{Appendix A. Omitted Proofs}

\label{sec:proofs}

In this section, we present the proofs of the main results introduced in Section~\ref{sec:theory}. We first introduce some useful lemmas in \hyperref[sec:useful-lemma]{A.1}, then prove the excess risk bounds given by Theorem~\ref{thm:uniform-allocation} and Theorem~\ref{thm:median-elimination} in \hyperref[sec:proof-theorem-ua]{A.2} and \hyperref[sec:proof-theorem-me]{A.3}. After that, we prove the exploratory regret bounds given by Lemma~\ref{lemma:ua_exploratory_regret} and Lemma~\ref{lemma:me_exploratory_regret} in \hyperref[sec:proof_of_lemma_ua_exploratory_regret]{A.4} and \hyperref[sec:proof_of_lemma_me_exploratory_regret]{A.5}, and finally we give the proof of one of the useful lemma originally proposed in this paper in \hyperref[sec:proof-useful-lemma]{A.6}. 

\subsection*{A.1.~Useful Lemmas}
\label{sec:useful-lemma}
We introduce two useful lemmas before the proof of main results. 

We first have the following lemma on the generalization error of the rejection model, which can be regarded as a counterpart of~\citep[Theorem 5]{ALT16:reject-theory}.

\begin{myLemma}
  \label{lemma:generalization-risk-0/1}
  Let $\mathcal{H}$ and $\mathcal{G}$ be the kernel-based hypotheses $\mathcal{H},\ \mathcal{G} = \{\x\mapsto\langle\w,\Phi(\x)\rangle\mid \Vert\w\Vert_{\mathbb{H}}\leq\Lambda\}$. Then for any $\delta>0$, with probability of $1-\delta$ over the draw of a sample $D$ of size $m$ from $\mathcal{D}$, the following holds for all $f \in\mathcal{H}\times\mathcal{G}$:
  \begin{equation}
  R(f)- \hat{R}_{D}^{surr}(f) \leq \frac{2-2\theta}{1-2\theta}\sqrt{\frac{(\kappa\Lambda)^2}{m}} + \sqrt{\frac{\log(1/\delta)}{2m}},
  \end{equation}
  where $\kappa^2 = \sup_{\x\in\X} K(\x,\x)$ and $K:\mathcal{X}\times\mathcal{X}\mapsto\mathbb{R}$ is the kernel function associated with $\mathbb{H}$.
\end{myLemma}

We then have the following lemma, which bounds the probability that a sub-optimal candidate feature is considered better than the optimal feature in a single empirical evaluation, which is the basic step in analyzing the effectiveness of feature exploration. The proof of Lemma~\ref{lemma:prob_ri_smaller_than_r1} can be found in Appendix~\hyperref[sec:proof-useful-lemma]{A.6}. 

\begin{myLemma}
  \label{lemma:prob_ri_smaller_than_r1}
  For any $i\in[K]$ with $\Delta_i>0$, if $\hat{f}_i$ is trained by ERM $\hat{R}_i^{surr}(f)$ on $n$ samples i.i.d. chosen in $\hat{D}_{tr, i}$, and $\hat{f}_1$ is trained by ERM $\hat{R}_1^{surr}(f)$ on $n$ samples i.i.d. chosen in $\hat{D}_{tr, 1}$, then 
  \begin{align*}
    \Pr\left[\hat{R}_i^{surr}(\hat{f}_i)<\hat{R}_1^{surr}(\hat{f}_1)\right]\leq&~4\exp\left(-\frac{2}{9}n\left(\frac{\Delta_i}{2}-\frac{2-2\theta}{1-2\theta}\sqrt{\frac{(\kappa\Lambda)^2}{n}}\right)^2\right),
  \end{align*}
  providing that the identification condition $n > \frac{16((1-\theta)\kappa\Lambda)^2}{((1-2\theta)\Delta)^2}$ holds, where $\Lambda=\sup_{i\in[K]}\Lambda_i$ ~and~ $\kappa=\sup_{i\in[K]}\sup_{\x\in\X_i}K_i(\x, \x)$. 
\end{myLemma}

\subsection*{A.2.~Proof of Theorem~\ref{thm:uniform-allocation}}
\label{sec:proof-theorem-ua}
\begin{proof}
  According to Eq.(\refeq{eq:key_decomposition}), the excess risk of learned model $\hat{f}_{i_s}$ can be decomposed into five parts,
  \begin{equation*}
    \begin{aligned}
      R_{i_s}(\hat{f}_{i_s}) - R_1^* &= \underbrace{R_{i_s}(\hat{f}_{i_s}) - \hat{R}_{tr, i_s}^{surr}(\hat{f}_{i_s})}_{\mathtt{term~(a)}} + \underbrace{\hat{R}_{tr, i_s}^{surr}(\hat{f}_{i_s}) - \hat{R}_{tr, 1}^{surr}(\hat{f}_1^*)}_{\mathtt{term~(b)}} \\
      &+ \underbrace{\hat{R}_{tr, 1}^{surr}(\hat{f}_1^*) - \hat{R}_{tr, 1}^{surr}(f_1^*)}_{\mathtt{term~(c)}} + \underbrace{\hat{R}_{tr, 1}^{surr}(f_1^*) - R_1^{surr}(f_1^*)}_{\mathtt{term~(d)}} + \underbrace{R_{ap}}_{\mathtt{term~(e)}},
    \end{aligned}
  \end{equation*}
  where $\mathtt{term~(a)}$ is the gap between the expected risk of the learned model $\hat{f}_{i_s}$ evaluated by $0/1$ loss and the empirical risk evaluated by surrogate loss, and $\mathtt{term~(b)}$ is the difference between empirical criterion of the selected feature and that of the best feature, where $\hat{f}_1^{*}$ refers to the best empirical model on the full training dataset augmented with best feature. Besides, $\mathtt{term~(c)}$ captures the difference between the empirical risk of $\hat{f}_1^{*}$ and that of the best hypothesis evaluated by surrogate loss $f_1^*=\argmin_{f\in\H_1\times\G_1}R_1^{surr}(f)$, $\mathtt{term~(d)}$ is the generalization error of $f_1^*$ evaluated by surrogate loss, and $\mathtt{term~(e)}$ is the unavoidable approximation error. Notice that $\mathtt{term~(c)}\leq 0$ by the definition of $\hat{f}_1^{*}$. Thus, to prove the theorem, it is sufficient to bound $\mathtt{term~(a)}$, $\mathtt{term~(b)}$ and $\mathtt{term~(d)}$ respectively.
  
  

  According to Lemma~\ref{lemma:generalization-risk-0/1}, for any $\delta_1>0$, $\mathtt{term~(a)}$ can be directly bounded by
  \begin{equation}
    \label{eq:term-a}
    \mathtt {term~(a)}\leq \frac{2-2\theta}{1-2\theta}\sqrt{\frac{(\kappa\Lambda)^2}{m}} + \sqrt{\frac{\log(1/\delta_1)}{2m}},
  \end{equation}
  with probability at least $1-\delta_1$. 
  
  By classical derivation of generalization bound based on Rademacher complexity, for any $\delta_2>0$, we have the following bound of $\mathtt{term~(d)}$ with probability at least $1-\delta_2$, 
  $$ \hat{R}_{tr, 1}^{surr}(f_1^*) - R_1^{surr}(f_1^*) \leq 2\mathfrak{R}_m(\tilde{\F}_1)+\sqrt{\frac{\log (1/\delta_2)}{2m}}, $$
  where $\tilde{\F}_1=\{\ell_{surr} \circ f~|~f\in\H_1\times\G_1\}$. According to~\citep[Theorem 5]{ALT16:reject-theory} we further have
  $$ \mathfrak{R}_m(\tilde{\F}_1) \leq \frac{1-\theta}{1-2\theta}\sqrt{\frac{(\kappa\Lambda)^2}{m}}, $$
  thus for any $\delta_2>0$ we obtain with probability at least $1-\delta_2$, 
  \begin{equation}
    \label{eq:term-d}
    \mathtt {term~(d)}\leq \frac{2-2\theta}{1-2\theta}\sqrt{\frac{(\kappa\Lambda)^2}{m}} + \sqrt{\frac{\log(1/\delta_2)}{2m}}.
  \end{equation}
  
  We then bound $\mathtt{term~(b)}$. By Lemma~\ref{lemma:ua_exploratory_regret}, for any $\delta_3>0$, we directly obtain with probability at least $1-\delta_3-\delta_{\text{fail}}$, 
  $$ \mathtt{term~(b)} \leq \frac{4-4\theta}{1-2\theta}\sqrt{\frac{(\kappa\Lambda)^2}{\lfloor B/K \rfloor}}+2\sqrt{\frac{\log(2/\delta_3)}{2\lfloor B/K \rfloor}}~.$$
  
  For any $\delta>0$, let $\delta_1=\delta_2=\delta_3=\delta/3$ and apply the union bound inequality, we have with probability at least $1-\delta-\delta_{\text{fail}}$, 
  \begin{align*}
    R_{i_s}(\hat{f}_{i_s})-R_1^* \leq {}&\frac{4-4\theta}{1-2\theta}\sqrt{\frac{(\kappa\Lambda)^2}{m}} + \frac{4-4\theta}{1-2\theta}\sqrt{\frac{(\kappa\Lambda)^2}{\lfloor B/K \rfloor}} + 2\sqrt{\frac{\log(3/\delta)}{2m}} + 2\sqrt{\frac{\log(6/\delta)}{2\lfloor B/K \rfloor}} + R_{ap} \\
    = {}& \O\left( \sqrt{\frac{(\kappa\Lambda)^2}{\lfloor B/K \rfloor}} + \sqrt{\frac{\log(6/\delta)}{2\lfloor B/K \rfloor}} \right) + R_{ap}.
  \end{align*}
  Finally, since $\lfloor B/K \rfloor > \frac{64((1-\theta)\kappa\Lambda)^2}{((1-2\theta)\Delta)^2}$, we have $\frac{\Delta}{2} - \frac{2-2\theta}{1-2\theta}\sqrt{\frac{(\kappa\Lambda)^2}{\lfloor B/K \rfloor}} \geq \frac{\Delta}{4}$, which is strictly greater than an absolute constant, so we can obtain an upper bound on fail probability as
  \begin{align*}
    \delta_{\text{fail}} &~= 4(K-1)\exp\left(-\frac{2}{9}\lfloor B/K \rfloor\left(\frac{\Delta}{2} - \frac{4-4\theta}{1-2\theta}\sqrt{\frac{(\kappa\Lambda)^2}{\lfloor B/K \rfloor}}\right)^2\right) \\
    &~\leq 4\left(K-1\right)\exp\left( -\frac{2}{9}\lfloor B/K \rfloor \frac{\Delta^2}{16}\right) \\
    &~= 4\left(K-1\right)\exp\left( -\frac{\Delta^2}{72}\lfloor B/K \rfloor\right)\\
    &~=\O\left( \exp\left(-\lfloor B/K \rfloor\right) \right),
  \end{align*}
  and the proof is finished.
\end{proof}

\subsection*{A.3.~Proof of Theorem~\ref{thm:median-elimination}}
\label{sec:proof-theorem-me}
\begin{proof}
  We first apply the same excess risk decomposition as shown in Eq.(\refeq{eq:key_decomposition}),
\begin{equation*}
  \begin{aligned}
    R_{i_s}(\hat{f}_{i_s}) - R_1^* &= \underbrace{R_{i_s}(\hat{f}_{i_s}) - \hat{R}_{tr, i_s}^{surr}(\hat{f}_{i_s})}_{\mathtt{term~(a)}} + \underbrace{\hat{R}_{tr, i_s}^{surr}(\hat{f}_{i_s}) - \hat{R}_{tr, 1}^{surr}(\hat{f}_1^*)}_{\mathtt{term~(b)}} \\
    &+ \underbrace{\hat{R}_{tr, 1}^{surr}(\hat{f}_1^*) - \hat{R}_{tr, 1}^{surr}(f_1^*)}_{\mathtt{term~(c)}} + \underbrace{\hat{R}_{tr, 1}^{surr}(f_1^*) - R_1^{surr}(f_1^*)}_{\mathtt{term~(d)}} + \underbrace{R_{ap}}_{\mathtt{term~(e)}} ,
  \end{aligned}
\end{equation*}
and $\mathtt{term~(a)}$, $\mathtt{term~(c)}$ and $\mathtt{term~(d)}$ can be bounded following the same derivation as in Theorem~\ref{thm:uniform-allocation}. According to Lemma~\ref{lemma:me_exploratory_regret}, we also have an upper bound of $\mathtt{term~(b)}$ with probability at least $1-\delta_3-\delta_{\text{fail}}$, 

$$\hat{R}_{tr, i_s}^{surr}(\hat{f}_{i_s})-\hat{R}_{tr, 1}^{surr}(\hat{f}_1^*)\leq \frac{4-4\theta}{1-2\theta}\sqrt{\frac{(\kappa\Lambda)^2}{\lfloor B/\log_2 K \rfloor}}+2\sqrt{\frac{\log(2/\delta_3)}{2\lfloor B/\log_2 K \rfloor}}, $$
where
$$\delta_{\text{fail}} =\frac{8\exp\left(-\frac{2}{9} \lfloor B/(K\log_2 K) \rfloor \left(\frac{\Delta}{2} - \frac{2-2\theta}{1-2\theta}\sqrt{\frac{(\kappa\Lambda)^2}{\lfloor B/(K\log_2 K) \rfloor}}\right)^2\right)}{1-\exp\left(-\frac{2}{9} \lfloor B/(K\log_2 K) \rfloor \left(\frac{\Delta}{2} - \frac{2-2\theta}{1-2\theta}\sqrt{\frac{(\kappa\Lambda)^2}{\lfloor B/(K\log_2 K) \rfloor}}\right)^2\right)}. $$

We then proceed to estimate the order of $\delta_{\text{fail}}$. Since $\lfloor B/(K\log_2 K) \rfloor > \frac{64((1-\theta)\kappa\Lambda)^2}{((1-2\theta)\Delta)^2}$ we have $\frac{\Delta}{2} - \frac{2-2\theta}{1-2\theta}\sqrt{\frac{(\kappa\Lambda)^2}{\lfloor B/K \rfloor}} \geq \frac{\Delta}{4}$, and so
\begin{align*}
  &~ \exp\left(-\frac{2}{9} \lfloor B/(K\log_2 K) \rfloor \left(\frac{\Delta}{2} - \frac{2-2\theta}{1-2\theta}\sqrt{\frac{(\kappa\Lambda)^2}{\lfloor B/(K\log_2 K) \rfloor}}\right)^2\right) \\
  \leq &~ \exp\left(-\frac{2}{9} \lfloor B/(K\log_2 K) \rfloor \frac{\Delta^2}{16}\right) \\
  = &~ \exp\left(-\frac{\Delta^2}{72} \lfloor B/(K\log_2 K) \rfloor \right),
\end{align*}
which upper-bounds the numerator of $\delta_{\text{fail}}$. Further let $ C=\frac{64((1-\theta)\kappa\Lambda)^2}{((1-2\theta)\Delta)^2} $ for simplicity as it appears to be a constant independent of $B$ and $K$. We conclude that 
\begin{align*}
  ~ & 1-\exp\left(-\frac{2}{9} \lfloor B/(K\log_2 K) \rfloor \left(\frac{\Delta}{2} - \frac{2-2\theta}{1-2\theta}\sqrt{\frac{(\kappa\Lambda)^2}{\lfloor B/(K\log_2 K) \rfloor}}\right)^2\right) \\
  \geq ~ & 1-\exp\left(-\frac{\Delta^2}{72} \lfloor B/(K\log_2 K) \rfloor \right) \\
  \geq ~ & 1-\exp\left( -\frac{\Delta^2C}{72} \right),
\end{align*}
which shows that the denominator of $\delta_{\text{fail}}$ is greater than an absolute constant independent of $B$, and so we have $\delta_{\text{fail}}=\O(\exp\left( -\lfloor B/(K\log_2 K) \rfloor \right))$. Again follow the derivation in the proof of Theorem~\ref{thm:uniform-allocation}, combine the results and set $\delta_1=\delta_2=\delta_3=\delta/3$ finishes the proof. 
\end{proof}

\subsection*{A.4.~Proof of Lemma~\ref{lemma:ua_exploratory_regret}}
\label{sec:proof_of_lemma_ua_exploratory_regret}
\begin{proof}
  If uniform allocation does not return the empirically best feature, then there must exists $i\in[K]$ s.t. $a_i$ is not the best feature, while its estimated risk is lower than the estimated risk of the best feature, i.e. $\hat{R}_i^{surr}(\hat{f}_i)<\hat{R}_1^{surr}(\hat{f}_1)$. Therefore, the algorithm returns the best feature with probability at least $1-\delta_{\text{fail}}$, where
\begin{align*}
  \delta_{\text{fail}} &= \Pr\left[\exists~i\in[K],~i\neq 1\wedge \hat{R}_i^{surr}(\hat{f}_i)<\hat{R}_1^{surr}(\hat{f}_1) \right] \\
  &\leq \sum_{i\in[K],i\neq 1}\Pr\left[\hat{R}_i^{surr}(\hat{f}_i)<\hat{R}_1^{surr}(\hat{f}_1)\right] \\
  &\leq 4\sum_{i\in[K],i\neq 1}\text{ }\exp\left(-\frac{2}{9}\lfloor B/K \rfloor\left(\frac{\Delta_i}{2}-\frac{2-2\theta}{1-2\theta}\sqrt{\frac{(\kappa\Lambda)^2}{\lfloor B/K \rfloor}}\right)^2\right) \\
  &\leq 4(K-1)\exp\left(-\frac{2}{9}\lfloor B/K \rfloor\left(\frac{\Delta}{2}-\frac{2-2\theta}{1-2\theta}\sqrt{\frac{(\kappa\Lambda)^2}{\lfloor B/K \rfloor}}\right)^2\right), 
\end{align*}
which proves the first part of the lemma. Specifically, the first inequality is because the union bound inequality, and the second inequality is according to Lemma~\ref{lemma:prob_ri_smaller_than_r1}.

To prove the second part, we firstly condition on the event that the algorithm has already identified an empirically best feature $a_1$. Define distribution $\mathcal{P}$ to be the uniform distribution on $\hat{D}_{tr, 1}$, we have
\begin{align*}
  &~\hat{R}_{tr, i_s}^{surr}(\hat{f}_{i_s})-\hat{R}_{tr, 1}^{surr}(\hat{f}_1^*) = \hat{R}_{tr, 1}^{surr}(\hat{f}_{1})-\hat{R}_{tr, 1}^{surr}(\hat{f}_1^*)\\
  =&~\underbrace{\hat{R}_{tr, 1}^{surr}(\hat{f}_1)-\hat{R}_1^{surr}(\hat{f}_1)}_{\mathtt{term~(a)}}+\underbrace{\hat{R}_1^{surr}(\hat{f}_1)-\hat{R}_1^{surr}(\hat{f}_1^*)}_{\mathtt{term~(b)}}+\underbrace{\hat{R}_1^{surr}(\hat{f}_1^*)-\hat{R}_{tr, 1}^{surr}(\hat{f}_1^*)}_{\mathtt{term~(c)}}
\end{align*}
where $\mathtt{term~(a)}$ is the generalization error of $\hat{f}_1$ on $\mathcal{P}$, $\mathtt{term~(b)}$ is the difference between the empirical error of the empirically best hypothesis $\hat{f}_1$ and the best hypothesis $\hat{f}_1^*$ on $\mathcal{P}$, and $\mathtt{term~(c)}$ is the generalization error of $\hat{f}_1^*$ on $\mathcal{P}$. Notice that $\mathtt{term~(b)}\leq 0$ since $\hat{f}_1$ minimizes $\hat{R}_1^{surr}$ by the ERM criterion. Thus, to prove the second part, it suffices to bound $\mathtt{term~(a)}$ and $\mathtt{term~(c)}$. By the standard analysis of generalization error based on the Rademacher complexity~\citep{book'2018:foundation}, with probability at least $1-\delta/2$, we have
\begin{align*}
  \mathtt{term~(a)} &\leq \frac{2-2\theta}{1-2\theta}\sqrt{\frac{(\kappa\Lambda)^2}{\lfloor B/K \rfloor}}+\sqrt{\frac{\log(2/\delta)}{2\lfloor B/K \rfloor}}~,
\end{align*}
and
  $$\mathtt{term~(c)}\leq \frac{2-2\theta}{1-2\theta}\sqrt{\frac{(\kappa\Lambda)^2}{\lfloor B/K \rfloor}}+\sqrt{\frac{\log(2/\delta)}{2\lfloor B/K \rfloor}}~.$$
Therefore, conditioning on the event that the algorithm returns a best feature, then with probability at least $1-\delta$, the uniform allocation algorithm satisfies
$$\hat{R}_{tr, i_s}^{surr}(\hat{f}_{i_s})-\hat{R}_{tr, 1}^{surr}(\hat{f}_1^*)\leq \frac{4-4\theta}{1-2\theta}\sqrt{\frac{(\kappa\Lambda)^2}{\lfloor B/K \rfloor}}+2\sqrt{\frac{\log(2/\delta)}{2\lfloor B/K \rfloor}}~.$$
Since the event occurs with probability at least $1-\delta_{\text{fail}}$, we conclude the second part of the proof by the union bound inequality. 
\end{proof}

\subsection*{A.5.~Proof of Lemma~\ref{lemma:me_exploratory_regret}}
\label{sec:proof_of_lemma_me_exploratory_regret}
\begin{proof}
Without loss of generality, assume throughout the proof that $K=2^c$ for some positive integer $c$, and that $B$ is a multiplier of $K\log_2K$. Let $n_t$ be the number of samples collected at round $t$, and $\hat{R}_{t, i}^{surr}(f)$ be the empirical surrogate risk on the samples collected at round $t$. Suppose $a_1\in\A_t$, and consider the probability $\delta_{\text{fail}}^{(t)}$ that $a_1$ is discarded at round $t$. For any $a_i\in\A_t$ s.t. $\Delta_i>0$, let $p_i^{(t)}$ be the probability that $\hat{R}_{t, i}^{surr}(\hat{f}_{t, i})<\hat{R}_{t, 1}^{surr}(\hat{f}_{t, 1})$ with $\hat{f}_{t, j}$ the models trained at round $t$. By Lemma~\ref{lemma:prob_ri_smaller_than_r1} we have
\begin{align*}
  p_i^{(t)} &\leq ~4\exp\left(-\frac{2}{9}n_t\left(\frac{\Delta_i}{2}-\frac{2-2\theta}{1-2\theta}\sqrt{\frac{(\kappa\Lambda)^2}{n_t}}\right)^2\right) \\
  &\leq ~4\exp\left(-\frac{2}{9}n_t\left(\frac{\Delta}{2}-\frac{2-2\theta}{1-2\theta}\sqrt{\frac{(\kappa\Lambda)^2}{n_1}}\right)^2\right)~.
\end{align*}
If $a_1$ is discarded at round $t$, then there must be at least $\frac{\abs{\A_t}}{2}$ features $a_i$ such that $a_i$ is not the best feature but $\hat{f}_{t, i}$ has lower empirical risk on $D_i$ than that of $\hat{f}_{t, 1}$ on $D_1$. Let $X$ be the random variable indicating the number of features satisfying the above property, it is easy to verify that 
$$\E\left[X\right]= \sum_{a_i\in\A_t}p_i^{(t)}\leq 4\abs{\A_t}\exp\left(-\frac{2}{9} n_t \left(\frac{\Delta}{2} - \frac{2-2\theta}{1-2\theta}\sqrt{\frac{(\kappa\Lambda)^2}{n_1}}\right)^2\right)~.$$ 
By Markov's inequality we have
$$\delta_{\text{fail}}^{(t)}=\Pr\left[X\geq \frac{\abs{\A_t}}{2}\right]\leq \frac{\E\left[X\right]}{\abs{\A_t}/2}\leq 8\exp\left(-\frac{2}{9} n_t \left(\frac{\Delta}{2} - \frac{2-2\theta}{1-2\theta}\sqrt{\frac{(\kappa\Lambda)^2}{n_1}}\right)^2\right)~.$$
So we can conclude that
$$\delta_{\text{fail}} = \sum_{t=1}^T\delta_{\text{fail}}^{(t)} \leq 8\sum_{t=1}^T\exp\left(-\frac{2}{9} n_t \left(\frac{\Delta}{2} - \frac{2-2\theta}{1-2\theta}\sqrt{\frac{(\kappa\Lambda)^2}{n_1}}\right)^2\right)~.$$
Since $\abs{\A_{t+1}}=\frac{\abs{\A_t}}{2}$, we have $n_{t+1}= 2n_t= \dots= 2^t n_1$. Therefore, 
\begin{align*}
  \delta_{\text{fail}} \leq & ~8\sum_{t=1}^T\exp\left(-\frac{2}{9} n_t \left(\frac{\Delta}{2} - \frac{2-2\theta}{1-2\theta}\sqrt{\frac{(\kappa\Lambda)^2}{n_1}}\right)^2\right) \\
  \leq & ~8\sum_{t=1}^T\exp\left(-\frac{2}{9} tn_1 \left(\frac{\Delta}{2} - \frac{2-2\theta}{1-2\theta}\sqrt{\frac{(\kappa\Lambda)^2}{n_1}}\right)^2\right) \\
  \leq & ~8\sum_{t=1}^\infty\exp\left(-\frac{2}{9} tn_1 \left(\frac{\Delta}{2} - \frac{2-2\theta}{1-2\theta}\sqrt{\frac{(\kappa\Lambda)^2}{n_1}}\right)^2\right) \\
  = &~ \frac{8\exp\left(-\frac{2}{9} n_1 \left(\frac{\Delta}{2} - \frac{2-2\theta}{1-2\theta}\sqrt{\frac{(\kappa\Lambda)^2}{n_1}}\right)^2\right)}{1-\exp\left(-\frac{2}{9} n_1 \left(\frac{\Delta}{2} - \frac{2-2\theta}{1-2\theta}\sqrt{\frac{(\kappa\Lambda)^2}{n_1}}\right)^2\right)},
\end{align*}
 which proves the first part of the lemma. 
 
 The second part shares the same derivation as that of Lemma~\ref{lemma:ua_exploratory_regret}, by which we have for any $\delta>0$, with probability at least $1-\delta-\delta_{\text{fail}}$, 
$$\hat{R}_{tr, i_s}^{surr}(\hat{f}_{i_s})-\hat{R}_{tr, 1}^{surr}(\hat{f}_1^*)\leq \frac{4-4\theta}{1-2\theta}\sqrt{\frac{(\kappa\Lambda)^2}{\sum_{t=1}^Tn_t}}+2\sqrt{\frac{\log (2/\delta)}{2\sum_{t=1}^Tn_t}}. $$
Finally, notice the fact that
\begin{align*}
  \sum_{t=1}^Tn_t &= n_1\sum_{t=1}^T2^t 
  =(2^{\lceil\log_2K\rceil+1}-1)\left\lfloor \frac{B}{K\log_2K} \right\rfloor 
\geq K\left\lfloor \frac{B}{K\log_2K} \right\rfloor 
=\O\left( \left\lfloor\frac{B}{\log_2K}\right\rfloor \right),
\end{align*}
which finishes the proof. 
\end{proof}

\subsection*{A.6.~Proof of Lemma~\ref{lemma:prob_ri_smaller_than_r1}}

\label{sec:proof-useful-lemma}

\begin{proof}[Proof of Lemma~\ref{lemma:prob_ri_smaller_than_r1}]
  If $\hat{R}_i^{surr}(\hat{f}_i)<\hat{R}_1^{surr}(\hat{f}_1)$, then it must be the case that either the estimation $\hat{R}_{i}^{surr}(\hat{f}_i)$ is over-optimistically, or the estimation $\hat{R}_{1}^{surr}(\hat{f}_1)$ is over-pessimistically. Let $p_i$ be the probability that $\hat{R}_i^{surr}(\hat{f}_i)<\hat{R}_1^{surr}(\hat{f}_1)$, then $p_i$ can be bounded by
\begin{align*}
  p_i &\leq \Pr\left[ \left(\hat{R}_{i}^{surr}(\hat{f}_i)<\hat{R}_{tr, i}^{surr}(\hat{f}_i^*)-\frac{\Delta_i}{2}\right)\vee \left(\hat{R}_{1}^{surr}(\hat{f}_1) > \hat{R}_{tr, 1}^{surr}(\hat{f}_1^*)+\frac{\Delta_i}{2}\right) \right] \\
  &\leq \underbrace{\Pr\left[ \hat{R}_{i}^{surr}(\hat{f}_i)<\hat{R}_{tr, i}^{surr}(\hat{f}_i^*)-\frac{\Delta_i}{2}\right]}_{\mathtt{term~(a)}} + \underbrace{\Pr\left[\hat{R}_{1}^{surr}(\hat{f}_1) > \hat{R}_{tr, 1}^{surr}(\hat{f}_1^*)+\frac{\Delta_i}{2} \right]}_{\mathtt{term~(b)}}~,
\end{align*}
where $\Delta_i=\hat{R}_{tr, i}^{surr}(\hat{f}_i^{*})-\min_{j\in[K]}\hat{R}_{tr, j}^{surr}(\hat{f}_j^{*})$ is defined in~\eqref{eq:optimality_gap}. We next explain how to upper bound ${\mathtt{term~(a)}}$, and the bound on ${\mathtt{term~(b)}}$ follows a similar derivation. First notice that
\begin{align*}
  \hat{R}_{tr, i}^{surr}(\hat{f}_i^*)-\hat{R}_i^{surr}(\hat{f}_i) &= \left(\hat{R}_{tr, i}^{surr}(\hat{f}_i^*)-\hat{R}_{tr, i}^{surr}(\hat{f}_i)\right)+\left(\hat{R}_{tr, i}^{surr}(\hat{f}_i)-\hat{R}_i^{surr}(\hat{f}_i)\right) \\
  &\leq \hat{R}_{tr, i}^{surr}(\hat{f}_i)-\hat{R}_i^{surr}(\hat{f}_i), 
\end{align*}
which is exactly a margin-based generalization bound. Define $\tilde{\mathcal{F}}_i=\left\{ \ell_{surr} \circ f~|~f\in\mathcal{H}_i\times\mathcal{G}_i \right\}$, standard generalization bound based on Rademacher complexity shows that for any $\delta>0$, with probability at least $1-\delta$, 
\begin{align*}
  \hat{R}_{tr, i}^{surr}(\hat{f}_i)-\hat{R}_i^{surr}(\hat{f}_i) & \leq 2\mathfrak{R}_m(\tilde{\mathcal{F}})+\sqrt{\frac{\log(1/\delta)}{2n}} \leq \frac{2-2\theta}{1-2\theta}\sqrt{\frac{(\kappa\Lambda)^2}{n}}+\sqrt{\frac{\log(1/\delta)}{2n}},
\end{align*}
where the second inequality is due to~\citep[Theorem 5]{ALT16:reject-theory}. Since $n > \frac{16((1-\theta)\kappa\Lambda)^2}{((1-2\theta)\Delta)^2}$, we have $\frac{\Delta_i}{2}\geq \frac{\Delta}{2}>\frac{2-2\theta}{1-2\theta}\sqrt{\frac{(\kappa\Lambda)^2}{n}}$, so the generalization error bound above can be translated as
\begin{align*}
  \Pr\left[ \hat{R}_{tr, i}^{surr}(\hat{f}_i)-\hat{R}_i^{surr}(\hat{f}_i) > \frac{\Delta_i}{2} \right] &\leq 2\exp\left(-\frac{2}{9}n \left(\frac{\Delta_i}{2}-\frac{2-2\theta}{1-2\theta}\sqrt{\frac{(\kappa\Lambda)^2}{n}}\right)^2\right).
\end{align*}
Since $\hat{R}_{tr, i}^{surr}(\hat{f}_i^*)-\hat{R}_i^{surr}(\hat{f}_i)\leq \hat{R}_{tr, i}^{surr}(\hat{f}_i)-\hat{R}_i^{surr}(\hat{f}_i)$, we conclude that
$$\mathtt{term~(a)}\leq 2\exp\left(-\frac{2}{9}n\left(\frac{\Delta_i}{2}-\frac{2-2\theta}{1-2\theta}\sqrt{\frac{(\kappa\Lambda)^2}{n}}\right)^2\right)~.$$
Similarly, we have
$$\mathtt{term~(b)}\leq 2\exp\left(-\frac{2}{9}n \left(\frac{\Delta_i}{2}-\frac{2-2\theta}{1-2\theta}\sqrt{\frac{(\kappa\Lambda)^2}{n}}\right)^2\right),$$
and the proof is finished.
\end{proof}

\end{document}